\documentclass[9.5pt,journal,final,finalsubmission,twocolumn]{IEEEtran}

\usepackage{amsmath}
\usepackage{amssymb}
\usepackage{subfigure, graphicx}
\usepackage[boxed]{algorithm2e}
\newtheorem{theorem}{Theorem}
\newtheorem{lemma}{Lemma}
\usepackage{epstopdf}

\begin{document}
%
\title{Non-Greedy L21-Norm Maximization for Principal Component Analysis}
%
%
\author{Feiping Nie, Heng Huang
\thanks{Feiping Nie and Heng Huang are with the Department of Computer Science and Engineering,
University of Texas at Arlington, USA. Email: feipingnie@gmail.com, heng@uta.edu}
}
\maketitle

\begin{abstract}
Principal Component Analysis (PCA) is one of the most important unsupervised methods to handle high-dimensional data. However, due to the high computational complexity of its eigen decomposition solution, it hard to apply PCA to the large-scale data with high dimensionality. Meanwhile, the squared L2-norm based objective makes it sensitive to data outliers. In recent research, the L1-norm maximization based PCA method was proposed for efficient computation and
being robust to outliers. However, this work used a greedy strategy to solve the eigen vectors. Moreover, the L1-norm maximization based objective may not be the correct robust PCA formulation, because it loses the theoretical connection to the minimization of data reconstruction error, which is one of the most important intuitions and goals of PCA. In this paper, we propose to maximize the L21-norm based robust PCA objective, which is theoretically connected to the minimization of reconstruction error. More importantly, we propose the efficient non-greedy optimization algorithms to solve our objective and the more general L21-norm maximization problem with theoretically guaranteed convergence. Experimental results on real world data sets show the effectiveness of the proposed method for principal component analysis.
\end{abstract}


\section{Introduction}

In many real-world applications, the dimensionality of data are very high. Directly handle the high-dimensional data is computationally expensive. At the same time, the performance could be poor because the number of available data is often limited and the noise in the data would increase dramatically when the dimensionality increases. To solve these problems, dimensionality reduction is one of the most important and effective methods. Among the dimensionality reduction algorithms, Principal Component Analysis (PCA) \cite{Jolliffe:pca} is one of the most widely used algorithms due to its simplicity and effectiveness. The main goal of PCA is to preserve the structure of the original data in the projected low-dimensional space. To this end, given a data set, PCA finds a projection matrix to minimize the reconstruction error of the projected data points under this projection matrix.

The traditional PCA has been successfully applied in many problems \cite{Duda:pattern} in the past decades. However, the traditional PCA algorithm has
several drawbacks. First, it need perform Singular Vector Decomposition (SVD) on input data matrix or eigen-decomposition on the covariance matrix, which is computationally expensive and difficult to be used when the number and the dimensionality of data are both very high. Second, it is sensitive to data outliers, because its objective function is intrinsically based on squared L2-norm and the outliers with large variation values can be exaggerated by the squared L2-norm. Many recent research works
\cite{rpca96,rpcaPAMI02,rpcaIJCV03,rpcaCVPR05,R1PCAicml06,rpcaNIPS09} have devoted effort to alleviate this problem and improve the robustness to outliers. \cite{rpca96,rpcaCVPR05} proposed to find the subspace such that the sum of L1-norm distances of data points to the subspace is minimized.
Although the robustness to outliers is improved in these methods, their algorithms are computationally expensive. Moreover, the used L1-norm in objective is not invariant to rotation. Thus, the performance is usually poor when their L1-norm based PCA is combined with $K$-means clustering \cite{R1PCAicml06}. To solve this problem, the R1-PCA was proposed with rotational invariant property and demonstrated good performance \cite{R1PCAicml06}. However, the R1-PCA iteratively performs the subspace iteration algorithm \cite{Golub:kyfan} in the high dimensional original space, which is computationally expensive. The extension of R1-PCA to tensor version can be found in \cite{rpcaCVPR08}.

Recently, a PCA method based on L1-norm maximization was proposed in \cite{RPCApami08}, and a similar work can be found in \cite{rpca87}. This method is invariant to rotation and is also robust to outliers. An efficient algorithm was proposed to solve the L1-norm maximization problem in \cite{RPCApami08}. This algorithm only need perform matrix-vector multiplication, and thus can be applied in the case that both the number and the dimensionality of data are very high. Several works on its tensor version and supervised version can be found in \cite{xuelongTSMCB10,rldaAAAI10,rpcaTCSVT10}. Due to the difficulty
of directly solving the L1-norm maximization problem, all these works use a greedy strategy to solve it. Specifically, the projection directions are optimized one by one sequentially. Such a kind of greedy method is easy to get stuck in a local solution.

Moreover, the L1-norm maximization based PCA method is not theoretically connected to minimization of the reconstruction error, which is the important goal of traditional PCA. In this paper, we propose a novel principal component analysis method based on the L21-norm maximization. The proposed method is robust to data outliers and also invariant to rotation. More importantly, our new method is theoretically connected to the minimization of reconstruction error, and thus is more suitable for principal component analysis than previous method in \cite{RPCApami08}. To solve the derived L21-norm PCA objective, we propose a new \emph{non-greedy} and efficient optimization algorithm to optimize all the projection directions simultaneously. Meanwhile, our algorithm will be extended to solve the more general maximization problems. We provide the theoretical analysis to guarantee the convergence of our algorithms. All experimental results on real world data sets show that the proposed method is effective for principal component analysis, and always obtains smaller reconstruction error than the method in \cite{RPCApami08} under the same reduced dimension.

The rest of this paper is organized as follows: We give a brief review of the related work in Section 2. In Section 3, we propose the L21-norm maximization based principal component analysis and solve the derived optimization problem through a new non-greedy and efficient algorithm which can solve the more general L21-norm maximization problem. In Section 4, we extend our algorithm to solve the general maximization problem which can be used to derive solutions for many other statistical learning models. In Section 5, we present experimental results to verify the effectiveness of the proposed method. Finally, we draw the conclusions in Section 6.

\section{Related work}

Given data $X = [x_1 ,x_2 ,\cdots,x_n ] \in \Re^{d \times n}$, where $d$ and $n$ are the dimensionality and number of data points respectively, without loss of generality, we can assume the data $\{x_i\}_{i=1}^n$ are centralized, \emph{i.e.}, $\sum_{i=1}^n x_i = 0$.

We denote the projection matrix $W = [w_1 ,w_2 ,\cdots,w_m ] \in \Re^{d \times m}$. The traditional PCA method minimizes the reconstruction error under the projected subspace, which is to solve the following optimization problem:
\begin{equation}
\label{opPCA1}
\mathop {\min }\limits_{W^T W = I} \sum\limits_{i = 1}^n {\left\| {x_i  - WW^T x_i } \right\|_2^2 },
\end{equation}
where $I$ is the identity matrix, $\| \cdot \|_2$ is the L2-norm of vector. Equivalently, the traditional PCA method can also be formulated as maximizing the
variance of data in the projected subspace, which is to solve the following optimization problem:
\begin{equation}
\label{opPCA}
 \mathop {\max }\limits_{W^T W = I} Tr(W^T S_t W) = \mathop {\max }\limits_{W^T W = I} \sum\limits_{i = 1}^n {\left\| {W^T x_i } \right\|_2^2 },
\end{equation}
where $S_t  = XX^T$ is the covariance matrix and $Tr(\cdot)$ is the trace operator of a matrix.
The equivalence of Eq.~(\ref{opPCA1}) and Eq.~(\ref{opPCA}) is based on the following equation for any matrix $W$ with $W^T W = I$:
\begin{equation}
\label{equal}
\sum\limits_{i = 1}^n {\left\| {x_i  - WW^T x_i } \right\|_2^2 }  + \sum\limits_{i = 1}^n {\left\| {W^T x_i } \right\|_2^2 }  = \sum\limits_{i = 1}^n {\left\| {x_i } \right\|_2^2 }
\end{equation}

Based on Eq.~(\ref{opPCA1}), R1-PCA was proposed to solve the following problem \cite{R1PCAicml06}:
\begin{equation}
\label{opR1PCA}
\mathop {\min }\limits_{W^T W = I} \sum\limits_{i = 1}^n {\left\| {x_i  - WW^T x_i } \right\|_2 }.
\end{equation}
R1-PCA minimizes the L2-norm loss instead of the squared L2-norm loss in traditional PCA, and thus the robustness to outliers is improved. The other important property of R1-PCA is that it is invariant to rotation, which is also preserved by traditional PCA.

Motivated by Eq.~(\ref{opPCA}), a recent work named PCA-L1 \cite{RPCApami08} was proposed to maximize the L1-norm instead of the squared L2-norm in traditional PCA by solving the following problem:
\begin{equation}
\label{opW}
\mathop {\max }\limits_{W^T W = I} \sum\limits_{i = 1}^n {\left\| {W^T x_i } \right\|_1 },
\end{equation}
where $\left\| \cdot \right\|_1$ is the L1-norm of vector. PCA-L1 also has the rotation invariant property. Directly solving this problem is difficult,
thus the author used a greedy strategy to solve it. Specifically, the $m$ projection directions $\{w_1 ,w_2 ,\cdots,w_m\}$ are optimized one by one. The first projection direction $w_1$ is optimized by solving:
\begin{equation}
  \mathop {\max }\limits_{w_1^T w_1 = 1} \sum\limits_{i = 1}^n {| {w_1^T x_i } |}.
\end{equation}

After the $(k-1)$-th projection direction $w_{k-1}$ has been obtained, the data matrix $X$ is transformed to $X = X - w_{k-1}(w_{k-1})^TX$, and then the $k$-th projection direction $w_k$ is optimized by solving:
\begin{equation}
\label{opgreed}
  \mathop {\max }\limits_{w_k^T w_k = 1} \sum\limits_{i = 1}^n {| {w_k^T x_i } |}.
\end{equation}

In this greedy method, the problem (\ref{opgreed}) is the key function to be solved for each $k$.
The work in \cite{RPCApami08} proposed an iterative algorithm to solve this problem. In order to guarantee the algorithm converges to a local maximum,
the algorithm adds an additional judgement after the convergence to $w_k^t$. If there exists $i$ such that $(w_k^t)^T x_i=0$, then let $w_k^t =
(w_k^t+\triangle w)/\|w_k^t+\triangle w\|_2$ and re-run the iterative algorithm, where $\triangle w$ is a small nonzero random vector. However, such
operation might make the algorithm interminable. For example, if there is a data point $x$ that exactly locates on the mean of the data set, then $x$ will be zero after centralization. As a result, $(w^t)^T x$ is always zero for any $w^t$). Moreover, it is possible that there exists $i$ such that $(w_k^t)^T x_i=0$ at the global maximum. In this case, the algorithm doesn't have the chance to find the global maximum.

\section{Principal component analysis with non-greedy L21-Norm maximization}

\subsection{L21-norm principal component analysis}
Motivated by Eq.~(\ref{opPCA}), we propose to solve the following problem:
\begin{equation}
\label{opPCA21}
\mathop {\max }\limits_{W^T W = I}  \sum\limits_{i = 1}^n {\| {W^T x_i } \|_2 } = \mathop {\max }\limits_{W^T W = I}  \| {X^T W } \|_{2,1},
\end{equation}
where $\left\| \cdot \right\|_{2,1}$ is the L21-norm of a matrix defined as $\left\| {M} \right\|_{2{\rm{,1}}}  =
\sum_i {( {\sum_j {m_{ij}^2 } })^{\frac{1}{2}} }$. 
Contrast to the name of PCA-L1 in \cite{RPCApami08} that solves problem (\ref{opW}), we call our PCA method with solving problem (\ref{opPCA21}) as PCA-L21. The PCA-L21 maximizes the L2-norm instead of the squared L2-norm in PCA, and thus the robustness to outliers is also improved. Similarly to R1-PCA and PCA-L1, PCA-L21 is also a rotation invariant method.

It is conjectured in \cite{RPCApami08} that problem (\ref{opR1PCA}) and problem (\ref{opW}) are closely related. However, no theoretical analysis was provided in \cite{RPCApami08} and it seems not the case according to our extensively experimental results. In contrast, we will show from both theoretical and experimental results that the proposed problem (\ref{opPCA21}) is indeed closely related to the problem (\ref{opR1PCA}), thus PCA-L21 is more suitable for the principal component analysis than PCA-L1.

First, we have the following lemma:
\begin{lemma}
\label{lem1}
  If $a^2  + b^2  = c^2$, then $\left| c \right| \le \left| a \right| + \left| b \right| \le \sqrt 2 \left| c \right|$.
\end{lemma}

\textbf{Proof}: Starting from the condition, we have
\begin{eqnarray}
\label{ineq01}
  &&a^2  + b^2  = c^2  \nonumber \\
  &\Rightarrow& a^2  + b^2  + 2\left| a \right|\left| b \right| \ge c^2 \nonumber \\
  &\Rightarrow& (\left| a \right| + \left| b \right|)^2 \ge c^2 \nonumber \\
  &\Rightarrow& \left| a \right| + \left| b \right| \ge \left| c \right|.
\end{eqnarray}
On the other hand, we have
\begin{eqnarray}
\label{ineq02}
&&(\left| a \right| - \left| b \right|)^2  \ge 0 \nonumber \\
&\Rightarrow& (\left| a \right| + \left| b \right|)^2  \le 2(a^2  + b^2 ) \nonumber \\
 &\Rightarrow& (\left| a \right| + \left| b \right|)^2  \le 2c^2  \nonumber \\
 &\Rightarrow& \left| a \right| + \left| b \right| \le \sqrt 2 \left| c \right|
\end{eqnarray}
Combining Eq.~(\ref{ineq01}) and Eq.~(\ref{ineq02}), we complete the proof.

\hfill $\Box$

According to Eq.~(\ref{equal}) and Lemma~\ref{lem1}, we have the following relationship:
\begin{equation}
  \sum\limits_{i = 1}^n {\left\| {x_i } \right\|_2 }  \le \sum\limits_{i = 1}^n {\left\| {x_i  - WW^T x_i } \right\|_2 }  + \sum\limits_{i = 1}^n {\left\| {W^T x_i } \right\|_2 }  \le \sqrt 2 \sum\limits_{i = 1}^n {\left\| {x_i } \right\|_2 },
\end{equation}
which can be written in matrix form as
\begin{equation}
\| X \|_{2,1}  \le  {\| {X^T - X^T WW^T} \|_{2,1} }  + \| {W^T X} \|_{2,1}  \le \sqrt 2 \left\| X \right\|_{2,1}.
\end{equation}

Therefore, the proposed problem (\ref{opPCA21}) is theoretically connected to the problem (\ref{opR1PCA}), which indicates that maximizing the
L21 norm as in problem (\ref{opPCA21}) also makes sense to minimize the reconstruction error, and thus is suitable for the
principal component analysis.

To solve the problem (\ref{opPCA21}), we first propose an efficient algorithm to solve the more general L21-norm maximization problem. Utilizing this
general algorithm, we can solve the problem (\ref{opPCA21}) directly without using the greedy strategy as in \cite{RPCApami08}.

\subsection{Efficient algorithm to solve the general L21-norm maximization problem}

Consider a general L21-norm maximization problem as follows:
\begin{equation}
\label{opl1norm}
  \mathop {\max }\limits_{v \in \mathcal{C}} f(v) + \sum\limits_i {\| {g_i (v)} \|_2},
\end{equation}
where $f(v)$ is an arbitrary scatter-output function, $g_i(v)$ (for each $i$) is an arbitrary vector-output function, and $v \in \mathcal{C}$ is an arbitrary constraint. We assume that the objective in problem (\ref{opl1norm}) has an upper bound.

We re-write the problem (\ref{opl1norm}) as the following problem:
\begin{equation}
\label{opreform}
\mathop {\max }\limits_{v \in \mathcal{C}} f(v) + \sum\limits_i {(\alpha _i)^T g_i (v)},
\end{equation}
where
\begin{equation}
\label{alpha}
\alpha _i  = \left\{ {\begin{array}{*{20}c}
   {\frac{{g_i (v)}}{{\left\| {g_i (v)} \right\|_2 }}} & {if \; \left\| {g_i (v)} \right\|_2  \ne 0}\ ;  \\
   \mathbf{0} & {if \; \left\| {g_i (v)} \right\|_2  = 0}\ .  \\
\end{array}} \right.
\end{equation}
Note that $\alpha _i$ depends on $v$ and thus is also an unknown variable. Based on Eqs.~(\ref{opreform}) and (\ref{alpha}), we propose an iterative algorithm to solve the problem (\ref{opl1norm}). The algorithm is described in Algorithm \ref{alg1}. In each iteration, $\alpha _i$ is updated by current solution $v$, and the solution $v$ is updated with the updated $\alpha _i$. The iterative procedure is repeated till the algorithm converges.

\vspace*{0pt}
\begin{algorithm}
\label{alg1}
Initialize $v^1 \in \mathcal{C}$, $t=1$ \;
\While{not converge}{
1. For each $i$, calculate $\alpha_i^t$ according to Eq.~(\ref{alpha}) \;
2. $v^{t+1} = \arg\mathop {\max }\limits_{v \in \mathcal{C}} f(v) + \sum\limits_i {(\alpha _i^t)^T g_i (v)}$ \;
3. $t=t+1$ \;
}
\caption{An efficient algorithm to solve the general L21-norm maximization problem (\ref{opl1norm}).}
\KwOut{$v^t$.}
\end{algorithm}
\vspace*{0pt}

Next, we prove that the proposed iterative algorithm will monotonically increase the objective function value of the problem (\ref{opl1norm}) in each iteration, and will converge to a local solution.

The convergence of the Algorithm \ref{alg1} is demonstrated in the following theorem:
\begin{theorem}
\label{converge}
The Algorithm \ref{alg1} monotonically increases the objective function value of the problem (\ref{opl1norm}) in each iteration.
\end{theorem}
\textbf{Proof}:
For each iteration $t$, according to the Step 2 in Algorithm \ref{alg1}, we have
\begin{equation}
\label{ineq1} f(v^{t + 1} ) + \sum\limits_i {(\alpha _i^t )^T g_i (v^{t + 1} )}
  \ge f(v^t ) + \sum\limits_i {(\alpha _i^t )^T g_i (v^t )}.
\end{equation}

On the other hand, for each $i$, according to the Cauchy-Schwarz inequality, we know:
\[\| {g_i (v^t )} \|_2 \| {g_i (v^{t + 1} )} \|_2  \ge g_i (v^t )^T g_i (v^{t + 1} ).\]

Based on this inequality, Eq.~(\ref{alpha}), and
\[\| {g_i (v^t )} \|_2  - (\alpha _i^t )^T g_i (v^t )=0,\]
we have
\begin{eqnarray}
&&\| {g_i (v^t )} \|_2 \| {g_i (v^{t + 1} )} \|_2  \ge g_i (v^t )^T g_i (v^{t + 1} ) \nonumber \\
&\Rightarrow& \| {g_i (v^{t + 1} )} \|_2  \ge (\alpha _i^t )^T g_i (v^{t + 1} ) \nonumber \\
&\Rightarrow& \| {g_i (v^{t + 1} )} \|_2  - (\alpha _i^t )^T g_i (v^{t + 1} ) \ge 0 \nonumber \\
&\Rightarrow& \| {g_i (v^{t + 1} )} \|_2  - (\alpha _i^t )^T g_i (v^{t + 1} ) \ge \| {g_i (v^t )} \|_2  - (\alpha _i^t )^T g_i (v^t ). \nonumber
\end{eqnarray}

The above inequality holds for every $i$, thus we have
\begin{equation}
\label{ineq2}
\begin{array}{l}
\sum\limits_i {\| {g_i (v^{t + 1} )} \|_2 }  - \sum\limits_i {(\alpha _i^t )^T g_i (v^{t + 1} )}  \\
\ge \sum\limits_i {\| {g_i (v^t )} \|_2 }  - \sum\limits_i {(\alpha _i^t )^T g_i (v^t )}\, .  \\
\end{array}
\end{equation}

Combining Eq.~(\ref{ineq1}) and Eq.~(\ref{ineq2}), we arrive at
\begin{equation}
f(v^{t + 1} ) + \sum\limits_i {\| {g_i (v^{t + 1} )} \|_2 }  \ge f(v^t ) + \sum\limits_i {\| {g_i (v^t )} \|_2 }\, .
\end{equation}

Thus, the Algorithm \ref{alg1} monotonically increases the objective of the problem (\ref{opl1norm}) in each iteration $t$.
\hfill $\Box$

As the objective of the problem (\ref{opl1norm}) has an upper bound, Theorem~\ref{converge} indicates that the Algorithm~\ref{alg1} converges. The following theorem shows that the Algorithm~\ref{alg1} will converge to a local solution.
\begin{theorem}
\label{local}
The Algorithm \ref{alg1} will  converge to a local solution of the problem~(\ref{opl1norm}).
\end{theorem}
\textbf{Proof}:
The Lagrangian function of the problem~(\ref{opl1norm}) is
\begin{equation}
\mathcal{L}(v,\lambda) = f(v) + \sum\limits_i {\| {g_i (v)} \|_2} -r(v,\lambda),
\end{equation}
where $r(\lambda,v)$ is the Lagrangian term to encode the constraint $v \in \mathcal{C}$ in problem~(\ref{opl1norm}).

Taking the derivative\footnote{When $g_i(v)=\mathbf{0}$, then $\mathbf{0}$ is a subgradient of function $\| g_i(v) \|_2$, so the $\alpha _i$ defined in Eq.~(\ref{alpha}) is the gradient or a subgradient of the function $\| g_i(v) \|_2$ in all the cases.} of $\mathcal{L}(v,\lambda)$ \emph{w.r.t.} $v$, and setting the derivative to zero, we have:
\begin{equation}
\label{kkt}
\frac{{\partial \mathcal{L}(v,\lambda)}}{{\partial v}} = \frac{{\partial f(v)}}{{\partial v}} + \sum\limits_i {{J_i(v) \alpha _i }} -
\frac{{\partial r(v,\lambda)}}{{\partial v}} = \mathbf{0},
\end{equation}
where $\alpha _i$ is defined in Eq.~(\ref{alpha}) and $J_i(v)$ is a matrix with the $(j,k)$-th element as $\frac{{\partial g_i^k(v)}}{{\partial v_j}}$, $g_i^k(v)$ denotes the $k$-th element of the vector $g_i(v)$.

Suppose the Algorithm \ref{alg1} converges to a solution $v^*$. From the Step 2 in Algorithm \ref{alg1} we have
\begin{equation}
\label{opKKT}
v^* = \arg\mathop {\max }\limits_{v \in \mathcal{C}} f(v^*) + \sum\limits_i {(\alpha_i^*)^T g_i (v^*)}.
\end{equation}

According to the KKT condition \cite{boydConvex} of the problem (\ref{opKKT}), we know that the solution $v^*$ for problem (\ref{opKKT}) satisfies Eq.~(\ref{kkt}). Note that Eq.~(\ref{kkt}) is the KKT condition of the problem (\ref{opl1norm}), hence the solution $v^*$ satisfies the KKT condition of the problem (\ref{opl1norm}). Therefore, the converged solution $v^*$ is a local solution of the problem (\ref{opl1norm}).

\hfill $\Box$

\subsection{Non-greedy maximization algorithm to solve L21-Norm Principal component analysis}

Obviously the proposed problem (\ref{opPCA21}) is a special case of the problem (\ref{opl1norm}), thus we can use the proposed Algorithm \ref{alg1} to solve the objective of L21-Norm PCA.

In Algorithm \ref{alg1}, Step 2 is the key step. Thus, to solve the problem (\ref{opPCA21}), the key step is to solve the following problem:
\begin{equation}
\label{opglobalstep}
\mathop {\max }\limits_{W^T W = I} \sum\limits_{i = 1}^n {\alpha _i^T W^T x_i }\,,
\end{equation}
where the vector $\alpha _i \in \Re^{m \times 1}$ is defined as:
\begin{equation}
\label{alpha1}
\alpha _i  = \left\{ {\begin{array}{*{20}c}
   {\frac{{W^T x_i}}{{\| {W^T x_i} \|_2 }}} & {if \; \| {W^T x_i} \|_2  \ne 0}\,;  \\
   \mathbf{0} & {if \; \| {W^T x_i} \|_2  = 0}\,.  \\
\end{array}} \right.
\end{equation}

Denoting the matrix $M = \sum_{i = 1}^n {x_i \alpha _i^T } \in \Re^{d \times m}$, we can re-write the problem (\ref{opglobalstep}) as:
\begin{equation}
\label{opglobal}
\mathop {\max }\limits_{W^T W = I} Tr(W^T M).
\end{equation}

Suppose the SVD result of $M$ is $M = U\Lambda V^T$, then $Tr(W^T M)$ can be re-written as:
\begin{eqnarray}
&&Tr(W^T M)  \nonumber \\
&=& Tr(W^T U\Lambda V^T )  \nonumber \\
&= & Tr(\Lambda V^T W^T U)  \nonumber \\
&=& Tr(\Lambda Z) \nonumber \\
&=&  \sum\limits_i {\lambda_{ii} z_{ii} }\,,
\end{eqnarray}
where $Z = V^T W^T U$, $\lambda_{ii}$ and $z_{ii}$ are the $(i,i)$-th element of matrix $\lambda$ and $Z$, respectively.

Note that $Z$ is an orthonormal matrix, {\em i.e.} $Z^T Z = I$, so $z_{ii} \le 1$. On the other hand, $\lambda_{ii} \ge 0$,  since $\lambda_{ii} $ is singular value of $M$. Therefore, $Tr(W^T M) = \sum\limits_i {\lambda_{ii} z_{ii} } \le \sum\limits_i {\lambda_{ii}}$, and when $z_{ii} = 1 \ (1\le i\le c)$, the equality holds. That is to say, $Tr(W^T M)$ reaches the maximum, when $Z=I$. Recall that $Z = V^T W^T U$, thus the optimal solution to the
problem Eq.~(\ref{opglobal}) is
\begin{equation}
\label{solQ}
  W = UZ^T V^T = UV^T.
\end{equation}

Based on the Algorithm \ref{alg1}, the algorithm of PCA-L21 to solve problem (\ref{opPCA21}) is described in Algorithm \ref{alg2}. According to Theorem \ref{local}, we can obtain a local solution with the algorithm. Contrast to the PCA-L1 algorithm in \cite{RPCApami08}, the PCA-L21 algorithm directly solves the projection matrix $W$ (\emph{i.e.} optimizes all projection directions simultaneously), but the PCA-L1 algorithm solves the projection directions one by one using a greedy strategy.

From Algorithm \ref{alg2} we can see that the computational complexity of the algorithm is $O(ndmt)$, where $n,d,m,t$ is the number of data points, the original dimensionality, the reduced dimensionality and the iteration number, respectively. In practice, the algorithm usually converges in
ten iterations. Therefore, the computational complexity of the algorithm is linear \emph{w.r.t.} either data number or data dimension, which indicates the algorithm is applicable in the case that both data number and data dimension are high. If the data are sparse, the computational complexity is further reduced to $O(nsmt)$, where $s$ is the averaged number of nonzero elements in one data point.

\vspace*{0pt}
\begin{algorithm}
\label{alg2}
\KwIn{$X$, $m$, where $X$ is centralized}
Initialize $W^1  \in \Re^{d \times m}$ such that $W^TW=I$, $t=1$ \;
\While{not converge}{
1. For each $i$, calculate $\alpha_i^t$ according to Eq.~(\ref{alpha1}), $M = \sum\limits_{i = 1}^n {x_i \alpha _i^T }$ \;
2. Calculate the SVD of $M$ as $M = U\Lambda V^T$, Let $W^{t+1} = UV^T$ \;
3. $t=t+1$ \;
}
\caption{The non-greedy optimization algorithm to solve the L21-norm principal component analysis.}
\KwOut{$W^t \in \Re^{d \times m}$.}
\end{algorithm}
\vspace*{0pt}

\subsection{Kernel and tensor extensions of L21-norm PCA}

Similar to traditional PCA, the proposed PCA-L21 is also a linear method, and is not suitable to handle data under the non-Gaussian distribution. A popular
technique to deal with this problem is to extend the linear method to kernel method. Obviously, the PCA-L21 is invariant to rotation and shift,
so this linear method satisfies the conditions of the general kernelization framework in \cite{generalKernel}. Thus, the PCA-L21 can be kernelized using this framework. Specifically, the given data are transformed by KPCA \cite{sch98KPCA}, and then Algorithm \ref{alg2} is performed on the transformed data.

Another problem of PCA is that it can only handle data points with vector format. For high-order tensor data, we have to vectorize the data to very high-dimensional vectors before applying PCA. This approach destroys the spacial information of tensor data and makes the computational burden very heavy.
A popular technique to deal with this problem is to extend the vector based method to tensor based method. As the problem (\ref{opPCA21}) in PCA-L21 only includes linear operator $W^Tx_i$, it can be easily extended to the tensor method to handle high-order data directly. For simplicity, we only briefly discuss the case of 2D tensor, the higher order tensor cases can be readily extended by replacing the linear operator $W^Tx_i$ with tensor operator \cite{phd:LATH97}.

Given data $X = [X_1 ,X_2 ,\cdots,X_n ] \in \Re^{r \times c \times n}$, where each data $X_i \in \Re^{r \times c}$ is a 2D matrix, $n$ is the number of data points, we assume that $\{X_i\}_{i=1}^n$ are centralized, \emph{i.e.}, $\sum_{i=1}^n X_i = \mathbf{0}$. To handle the tensor case, linear operator $W^Tx_i$ is replaced by $U^T X_i V$, where $U \in \Re^{r \times k_1}$ and $V \in \Re^{c \times k_2}$ are two projection matrices ($k_1<r$ and $k_2<c$ are the reduced dimensions of two projection subspaces). Correspondingly, the problem (\ref{opPCA21}) becomes:
\begin{equation}
\label{optensor}
\mathop {\max }\limits_{U^T U = I_{k_1 } ,V^T V = I_{k_2} } \sum\limits_{i = 1}^n {\| {U^T X_i V} \|_F },
\end{equation}
where $\| \cdot \|_F$ is the Frobenius norm of matrix.
Similar to other tensor methods, problem (\ref{optensor}) can be solved by alternative optimization technique. Specifically, when fix $U$, the problem (\ref{optensor}) is reduced to the problem (\ref{opPCA21}), and thus the $V$ can be optimized by Algorithm \ref{alg2}. In turn, $U$ can also be optimized by Algorithm \ref{alg2} when fix $V$. The procedure is iteratively performed until converges.

\section{The Extension of our algorithm for general maximization problem}

Besides solving the L21-norm maximization problem in above section, to provide the useful and efficient algorithm for related research problems, we extend our idea to solve the more general maximization problem as follows:
\begin{equation}
\label{opl10norm}
  \mathop {\max }\limits_{v \in \mathcal{C}} f(v) + \sum\limits_i {h_i (g_i(v))},
\end{equation}
where $f(v)$ is an arbitrary scatter-output function, $g_i(v)$ (for each $i$) is an arbitrary scatter, vector, or matrix-output function, and $h_i$ (for each $i$) is an arbitrary \textbf{convex} function, $v \in \mathcal{C}$ is an arbitrary constraint. We assume that the objective in problem (\ref{opl10norm}) has an upper bound.

We propose an iterative algorithm to solve the problem (\ref{opl1norm}). The algorithm is described in Algorithm \ref{alg10}. Similar to Algorithm \ref{alg1}, in each iteration, $\alpha _i$ is updated by current solution $v$, and the solution $v$ is updated with the updated $\alpha _i$. The iterative procedure is repeated till the algorithm converges.

\vspace*{0pt}
\begin{algorithm}
\label{alg10}
Initialize $v^1 \in \mathcal{C}$, $t=1$ \;
\While{not converge}{
1. For each $i$, calculate $\alpha_i^t = h_i'(g_i(v^t)) = \frac{{\partial h_i(g_i(v^t))}}{{\partial (g_i(v^t))}}$ \;
2. $v^{t+1} = \arg\mathop {\max }\limits_{v \in \mathcal{C}} f(v) + \sum\limits_i {Tr((\alpha _i^t)^T g_i (v))}$ \;
3. $t=t+1$ \;
}
\caption{An efficient algorithm to solve the more general maximization problem (\ref{opl10norm}).}
\KwOut{$v^t$.}
\end{algorithm}
\vspace*{0pt}

The convergence of the Algorithm~\ref{alg10} is guaranteed by the following theorem:
\begin{theorem}
\label{converge0}
The Algorithm \ref{alg10} monotonically increases the objective of the problem (\ref{opl10norm}) in each iteration.
\end{theorem}
\textbf{Proof}:
For each iteration $t$, according to the Step 2 in Algorithm~\ref{alg10}, we have
\begin{equation}
\label{ineq10}
\begin{array}{l}
f(v^{t + 1} ) + \sum\limits_i {Tr((\alpha _i^t )^T g_i (v^{t + 1} ))}  \ge f(v^t ) + \sum\limits_i {Tr((\alpha _i^t )^T g_i (v^t ))}.
 \end{array}
\end{equation}

For each $i$, since $h_i$ is convex, according to the property of convex function, we know $h_i (g_i (v^{t + 1} )) - h_i (g_i (v^t )) \ge Tr((g_i (v^{t + 1} ) - g_i (v^t ))^T h'_i (g_i (v^t ))$. According to the definition of $\alpha _i^t$ in Step 1, we have
\begin{equation}
\label{ineq20}
\begin{array}{l}
 h_i (g_i (v^{t + 1} )) - Tr((\alpha _i^t )^T g_i (v^{t + 1} )) \\
  \ge h_i (g_i (v^t )) - Tr((\alpha _i^t )^T g_i (v^t )).
 \end{array}
\end{equation}

Combining Eq.~(\ref{ineq10}) and Eq.~(\ref{ineq20}), we arrive at
\begin{equation}
f(v^{t + 1} ) + \sum\limits_i {h_i (g_i (v^{t + 1} ))}  \ge f(v^t ) + \sum\limits_i {h_i (g_i (v^t ))}.
\end{equation}
Thus the Algorithm \ref{alg10} monotonically increases the objective of the problem (\ref{opl10norm}) in each iteration.

\hfill $\Box$

Because the objective of the problem (\ref{opl10norm}) has an upper bound, Theorem \ref{converge0} indicates that the Algorithm \ref{alg10} converges. The following theorem shows that the Algorithm \ref{alg10} will converge to a local solution of the problem~(\ref{opl10norm}).

\begin{theorem}
\label{local0}
The Algorithm \ref{alg10} will  converge to a local solution of the problem~(\ref{opl10norm}).
\end{theorem}
\textbf{Proof}:
The Lagrangian function of the problem~(\ref{opl10norm}) is
\begin{equation}
\mathcal{L}(v,\lambda) = f(v) + \sum\limits_i {h_i (g_i(v))} -r(v,\lambda),
\end{equation}
where $r(\lambda,v)$ is the Lagrangian term to encode the constraint $v \in \mathcal{C}$ in problem~(\ref{opl10norm}).

Taking the derivative
\footnote{When $x=\mathbf{0}$, then $\mathbf{0}$ is a subgradient of function $\| x \|_2$, so $\alpha _i$ is the gradient or a subgradient of the function $\| x \|_2$ in all the cases, where $\alpha _i  = {{{x}}/{{\| {x} \|_2 }}}$
if $\| {x} \|_2  \ne 0$ and $\alpha _i=\mathbf{0}$ otherwise.}
of $\mathcal{L}(v,\lambda)$ \emph{w.r.t.} $v$, and setting the derivative to zero, we have:
\begin{equation}
\label{kkt0}
\frac{{\partial \mathcal{L}(v,\lambda)}}{{\partial v}} = \frac{{\partial f(v)}}{{\partial v}} + \sum\limits_i { \frac{{Tr\left( {\left( \alpha _i \right)^T \partial g_i(v)} \right)}}{{\partial v}} } -
\frac{{\partial r(v,\lambda)}}{{\partial v}} = \mathbf{0},
\end{equation}
where $\alpha _i=h_i'(g_i(v))$.

Suppose the Algorithm \ref{alg10} converges to a solution $v^*$. From the Step 2 in Algorithm \ref{alg10} we have
\begin{equation}
\label{opKKT0}
v^* = \arg\mathop {\max }\limits_{v \in \mathcal{C}} f(v^*) + \sum\limits_i {Tr((\alpha_i^*)^T g_i (v^*))}.
\end{equation}

According to the KKT condition of the problem (\ref{opKKT0}), we know that
the solution $v^*$ to problem (\ref{opKKT0}) satisfies Eq.~(\ref{kkt0}). Note that Eq.~(\ref{kkt0}) is the KKT
condition of the problem (\ref{opl10norm}), so the converged solution $v^*$ of Algorithm \ref{alg10} satisfies the KKT condition of the problem (\ref{opl10norm}).
Therefore, the converged solution $v^*$ is a local solution of the problem (\ref{opl10norm}).
\hfill $\Box$

Algorithm \ref{alg10} is very useful to solve the general maximization problems. For example, we can directly use the algorithm to solve
the following two important problems:
\begin{equation}
\label{op_pm}
\mathop {\max }\limits_{V^T V = I} Tr(V^T AV) \, ,
\end{equation}
\begin{equation}
\label{op_tpm}
\mathop {\max }\limits_{v^T v = 1,\left\| v \right\|_{0}  \le k} v^T Av \, .
\end{equation}

It is interesting to point out that the derived algorithms for Eq.~(\ref{op_pm}) and Eq.~(\ref{op_tpm}) based on Algorithm~\ref{alg10} are exactly the classical power method (or subspace iteration method) and the recently proposed truncated power method \cite{powermethod11}, respectively.

\section{Experiments}

In this section, we will present the experimental results to demonstrate the effectiveness of the proposed PCA-L21 compared to
traditional PCA, R1-PCA and PCA-L1.

\subsection{Reconstruction errors with occlusions}

\begin{figure}
\vskip 0.in
  \centering
\includegraphics[height=4.5cm]{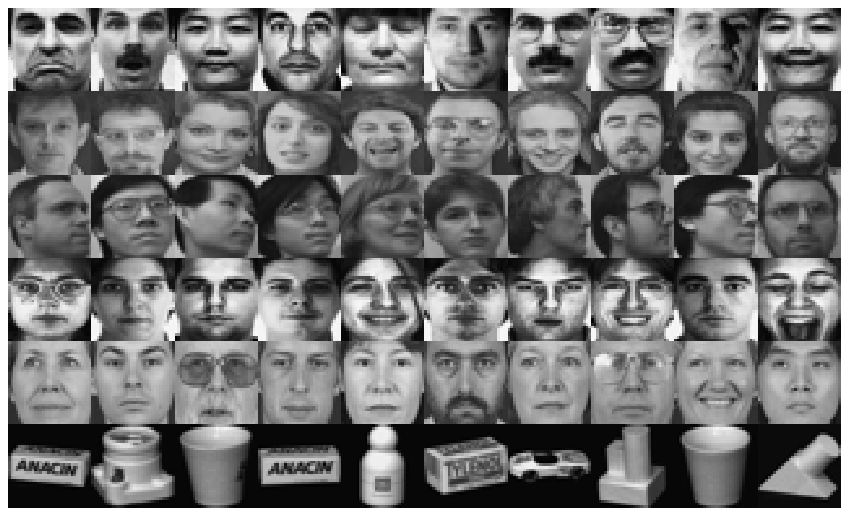}
\vskip -0.in
  \caption{Some samples from six benchmark data sets. The images from the first row to the sixth row are: Yale, AT\&T, Umist, AR, XM2VTS, Coil20.}
\vskip -0.in
\label{datasamples}
\end{figure}

We use six image benchmark data sets to perform our experiments. A brief descriptions of the data sets are shown
in Table \ref{tabData}, and the samples from each data sets are shown in Figure \ref{datasamples}.

\begin{table}[t]
\caption{Data Descriptions.}
 \vskip -0in
\begin{center} \small
\begin{tabular}{|c|c|c|}
\hline
Data set & Number of data & Dimensionality \\
\hline
Yale & 165 & 3456 \\
 \hline
AT\&T & 400 & 2576 \\
 \hline
Umist & 575 & 2576 \\
 \hline
AR & 840 & 3072 \\
 \hline
XM2VTS & 1180 & 4096 \\
 \hline
Coil20 & 1440 & 4096 \\
 \hline
\end{tabular}
\end{center}
\label{tabData}
\end{table}

In each image data set, we randomly select 10, 20, 30 percent images respectively, and each selected image is occluded with a randomly located square.
The width of these squares are the half of the width of the images.

We use the following reconstruction error to measure the quality of dimensionality reduction methods:
\begin{equation}
\label{rce}
e(m) = \frac{1}{n} \sum\limits_{i = 1}^n {\left\| {x_i^{org}  - WW^T x_i } \right\|_2 },
\end{equation}
where $n$ is the number of training data, $W \in \Re^{d \times m}$ is the learned projection matrix by PCA, R1-PCA, PCA-L1, or PCA-L21,
$x_i^{org}$ and $x_i$ are the $i$-th original non-occluded image and the $i$-th image used in the
training respectively.

\begin{figure*}[!t]
\vskip 0.in
  \centering
  \subfigure[Yale]{
    \label{ionosphere} 
    \includegraphics[height=4.cm]{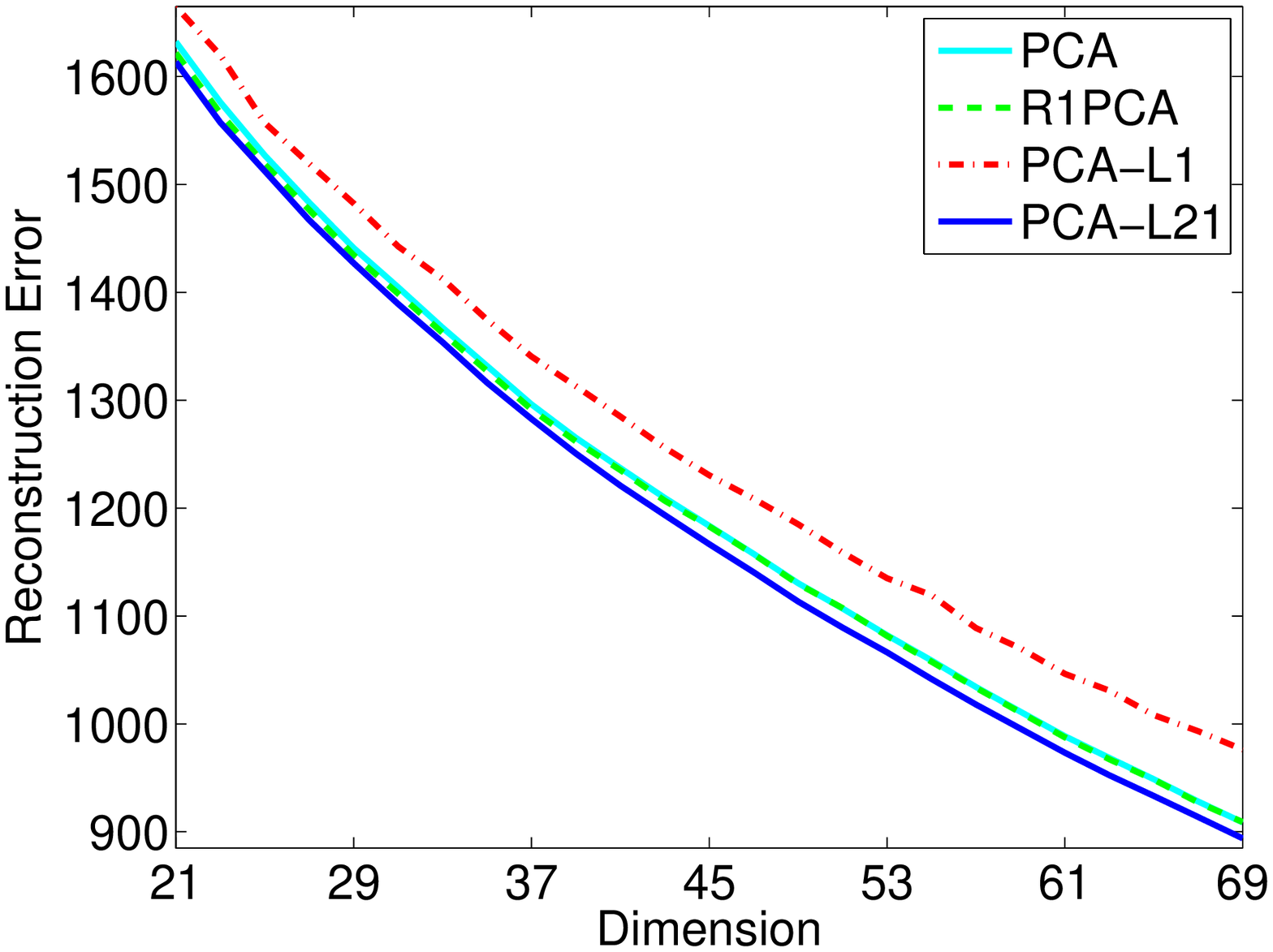}}
  \subfigure[AT\&T]{
    \label{waveform} 
    \includegraphics[height=4.cm]{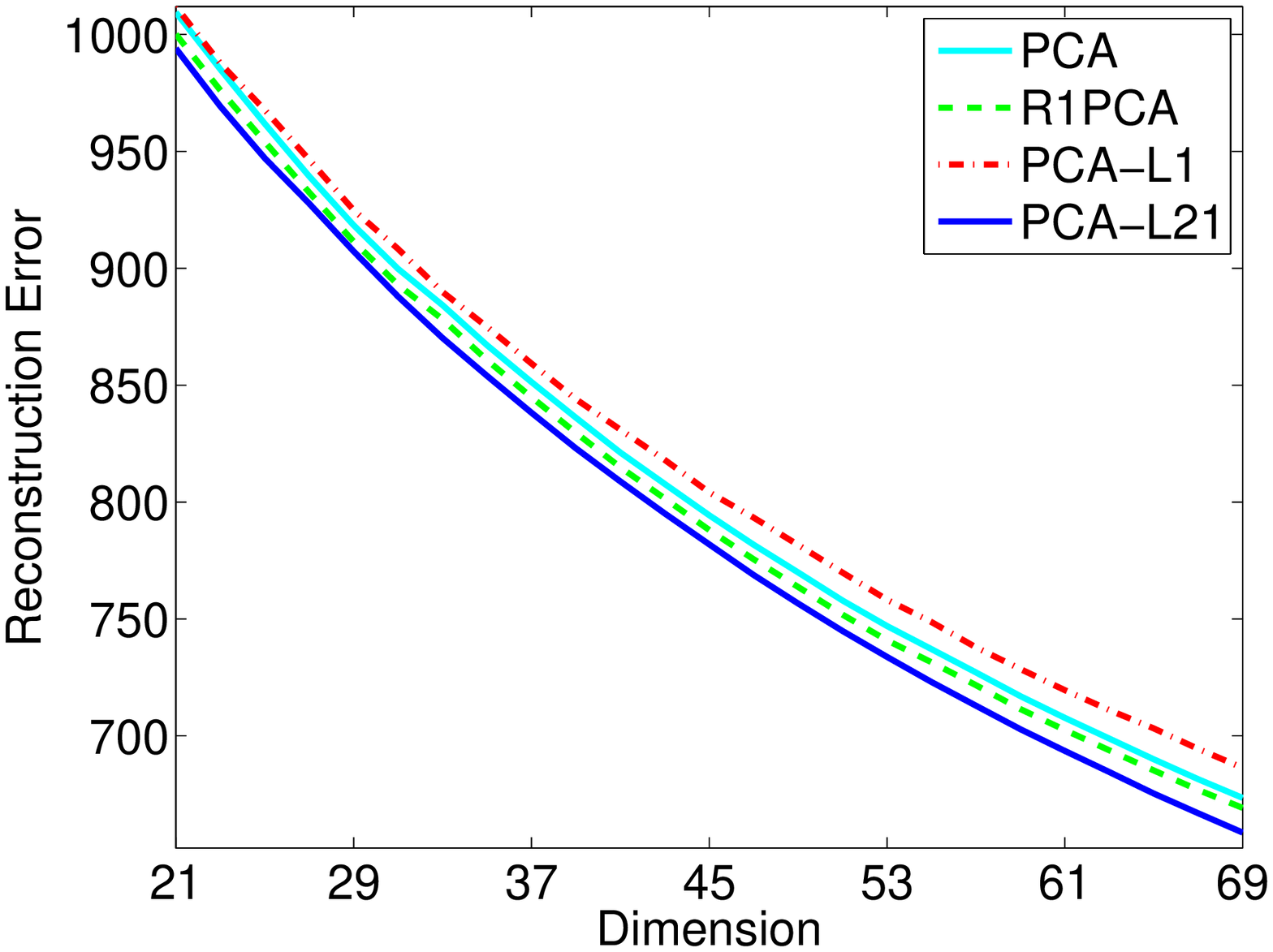}}
  \subfigure[Umist]{
    \label{waveform} 
    \includegraphics[height=4.cm]{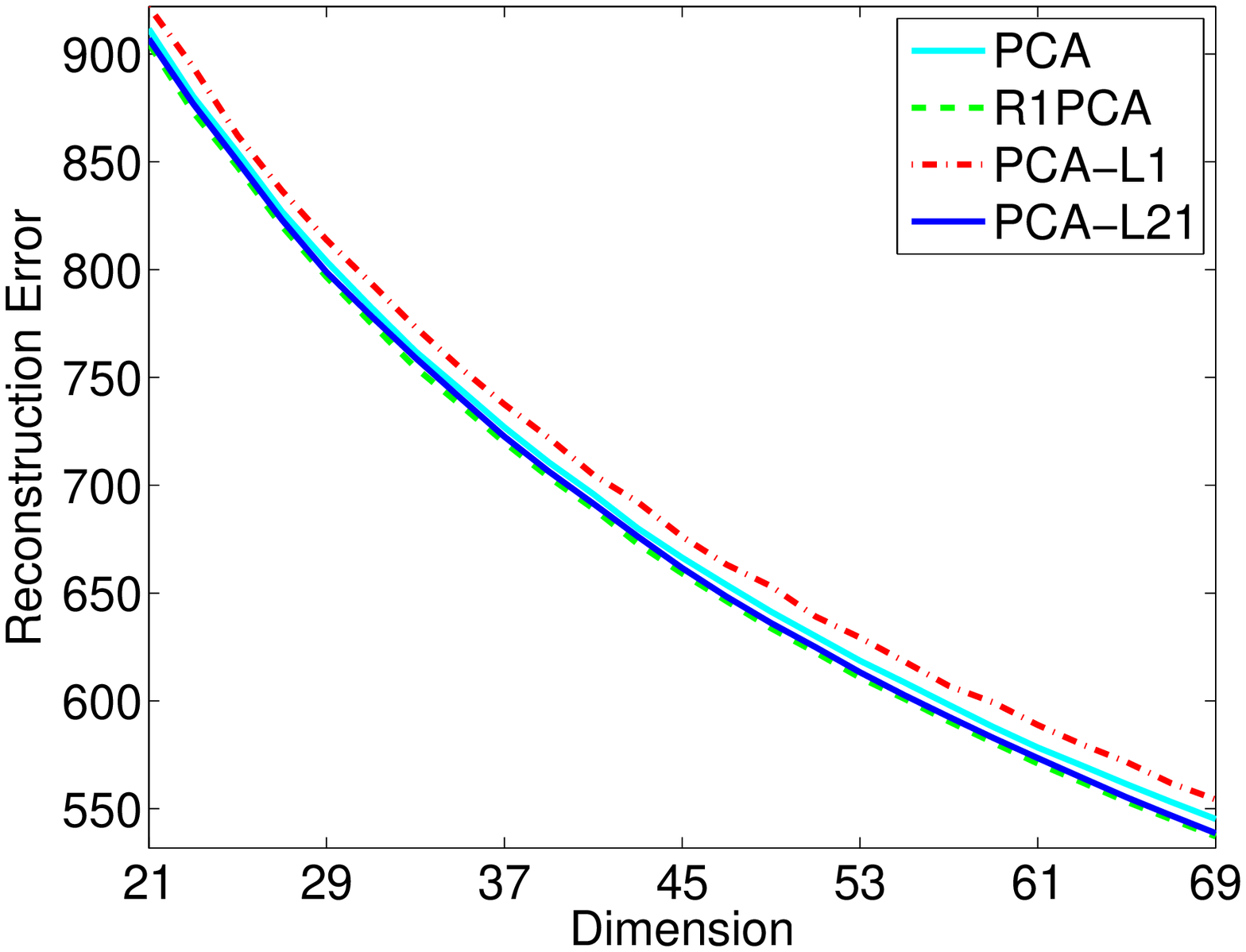}}
  \subfigure[AR]{
    \label{waveform} 
    \includegraphics[height=4.cm]{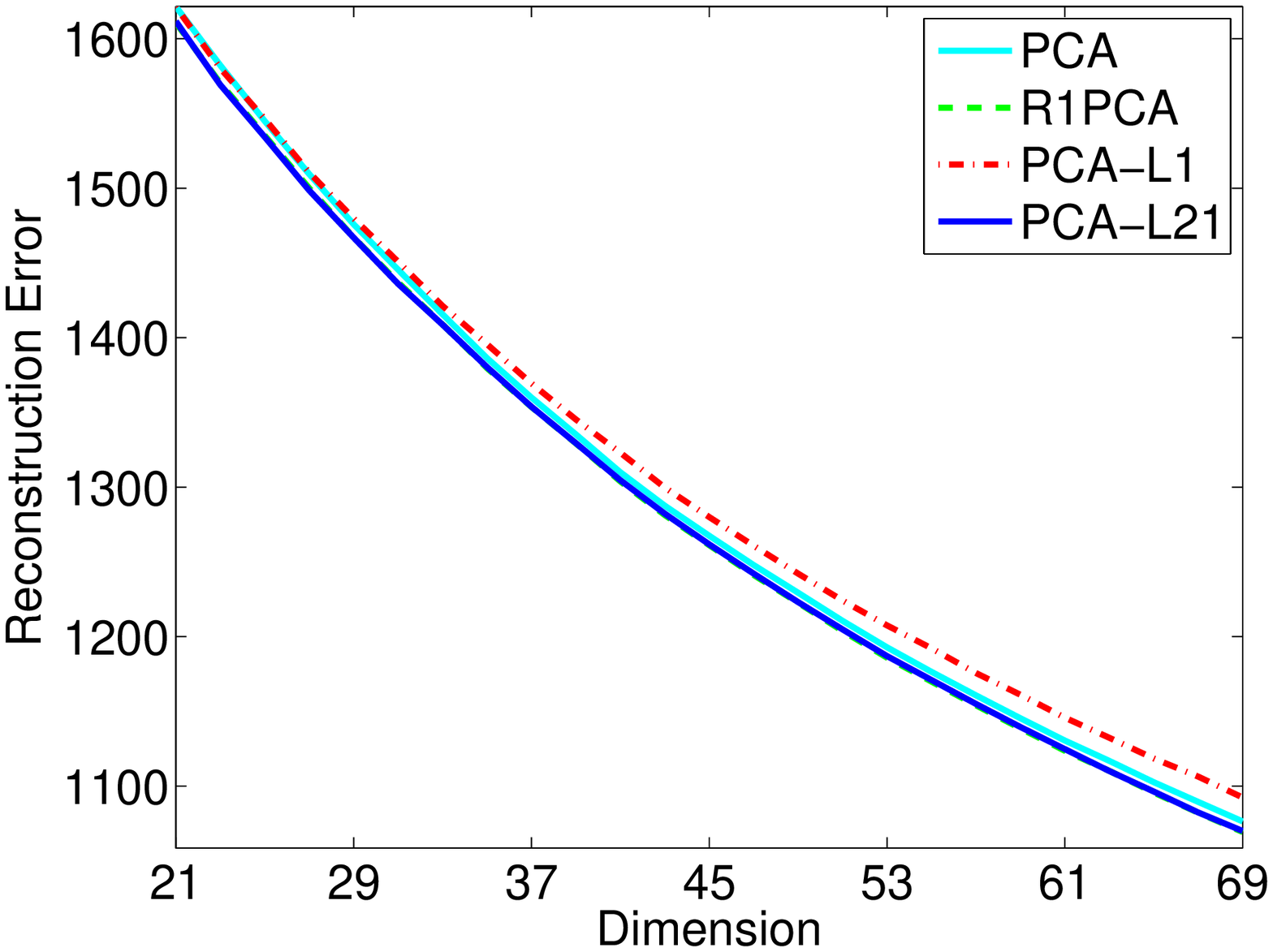}}
  \subfigure[XM2VTS]{
    \label{waveform} 
    \includegraphics[height=4.cm]{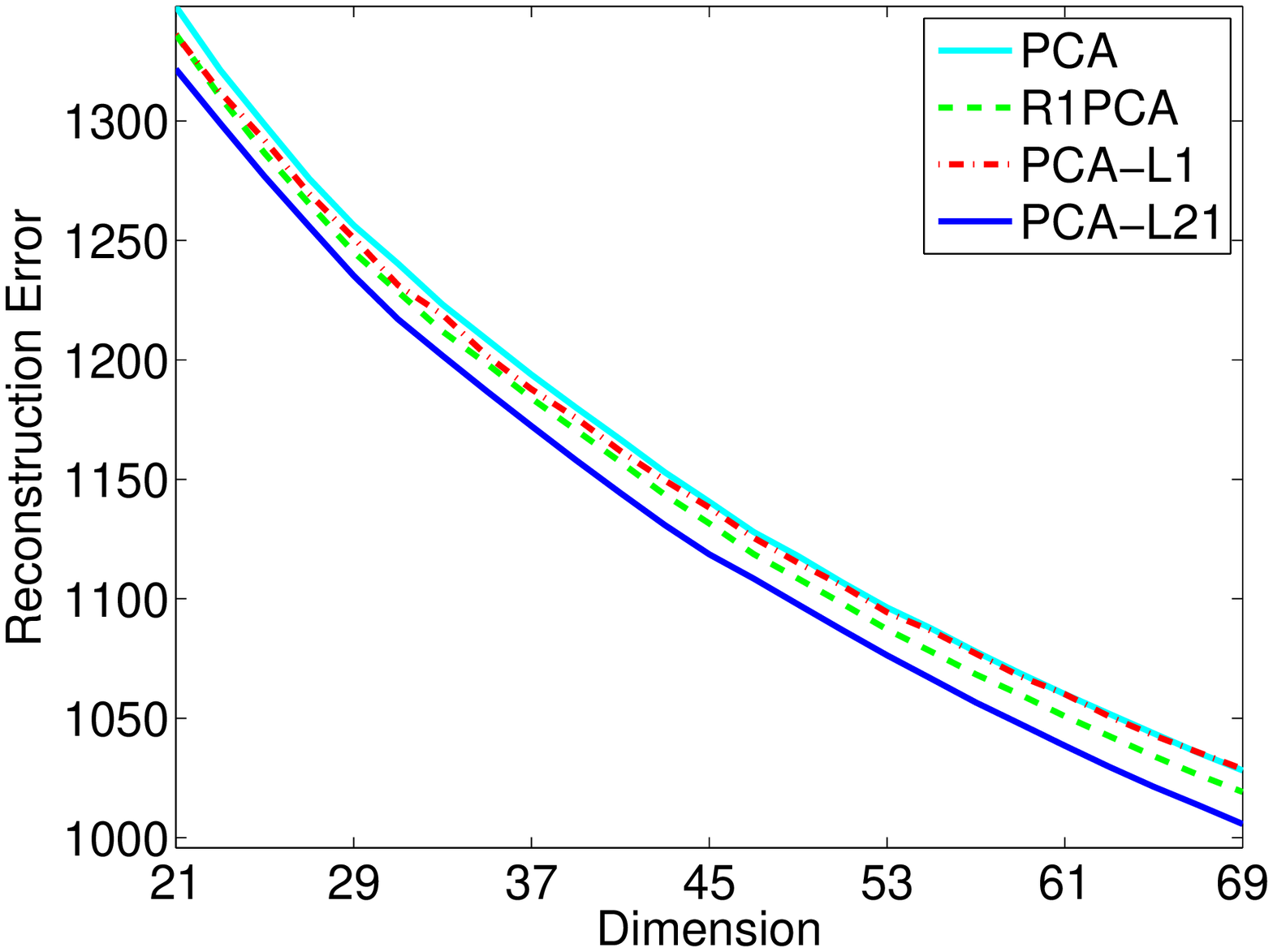}}
  \subfigure[Coil20]{
    \label{waveform} 
    \includegraphics[height=4.cm]{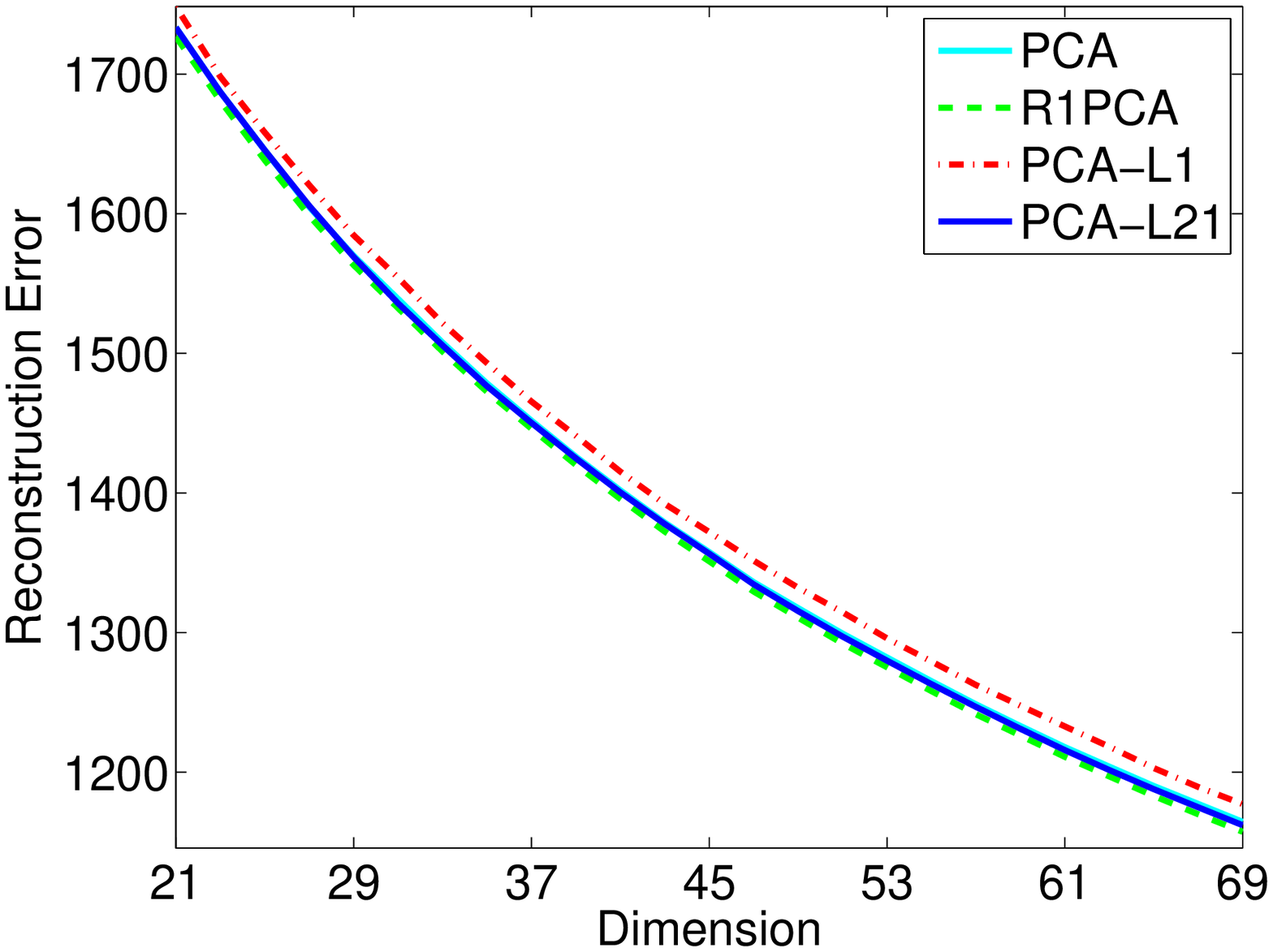}}
  \caption{Reconstruction errors (calculated by Eq.~(\ref{rce})) under different dimensions obtained by PCA, R1PCA, PCA-L1 and PCA-L21, respectively.
  In each data set, 10 percent images are randomly occluded with a randomly located square. }
  \label{objective1} 
\vskip 0.in
\end{figure*}

\begin{figure*}[!t]
\vskip 0.in
  \centering
  \subfigure[Yale]{
    \label{ionosphere} 
    \includegraphics[height=4.cm]{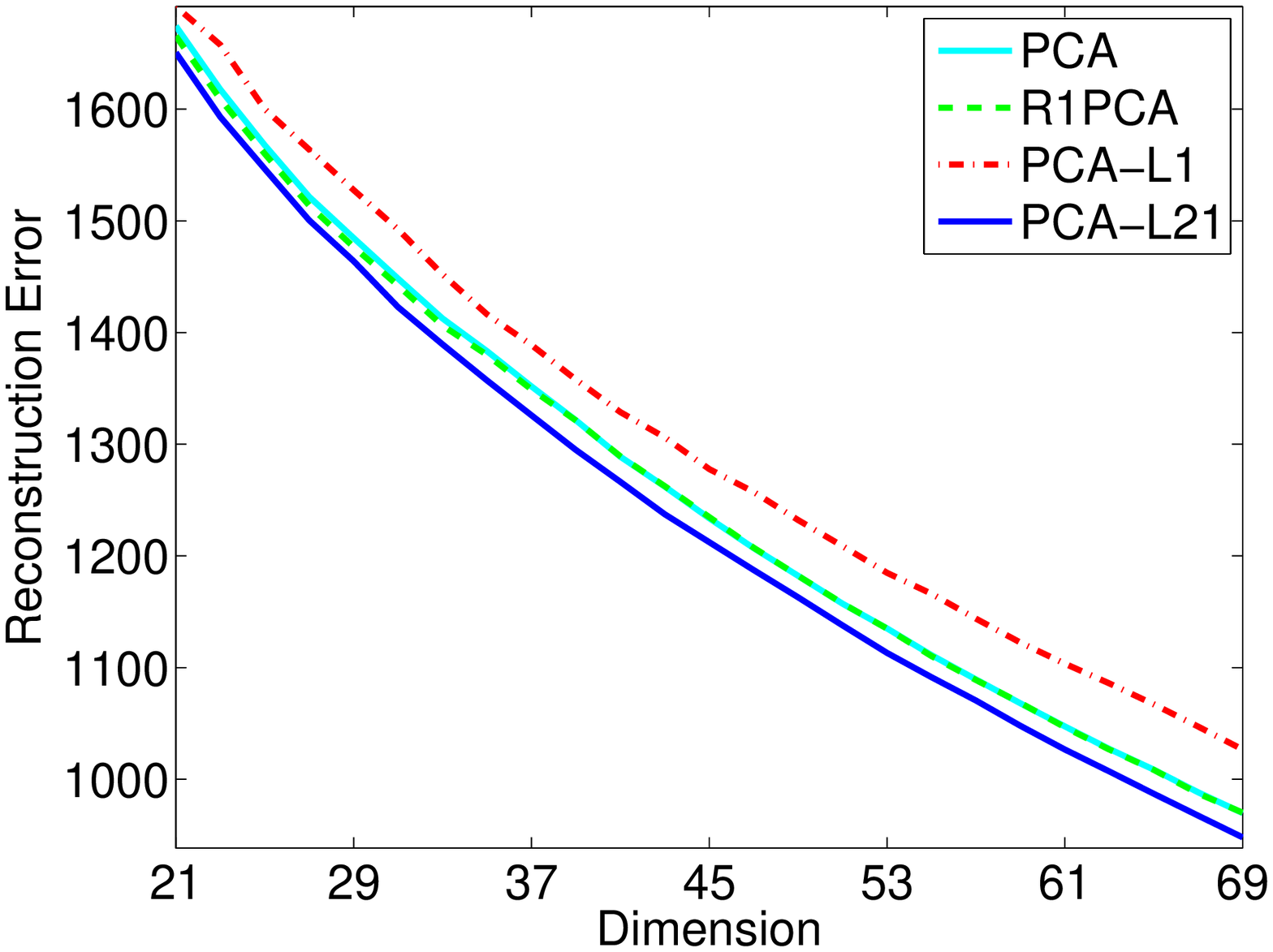}}
  \subfigure[AT\&T]{
    \label{waveform} 
    \includegraphics[height=4.cm]{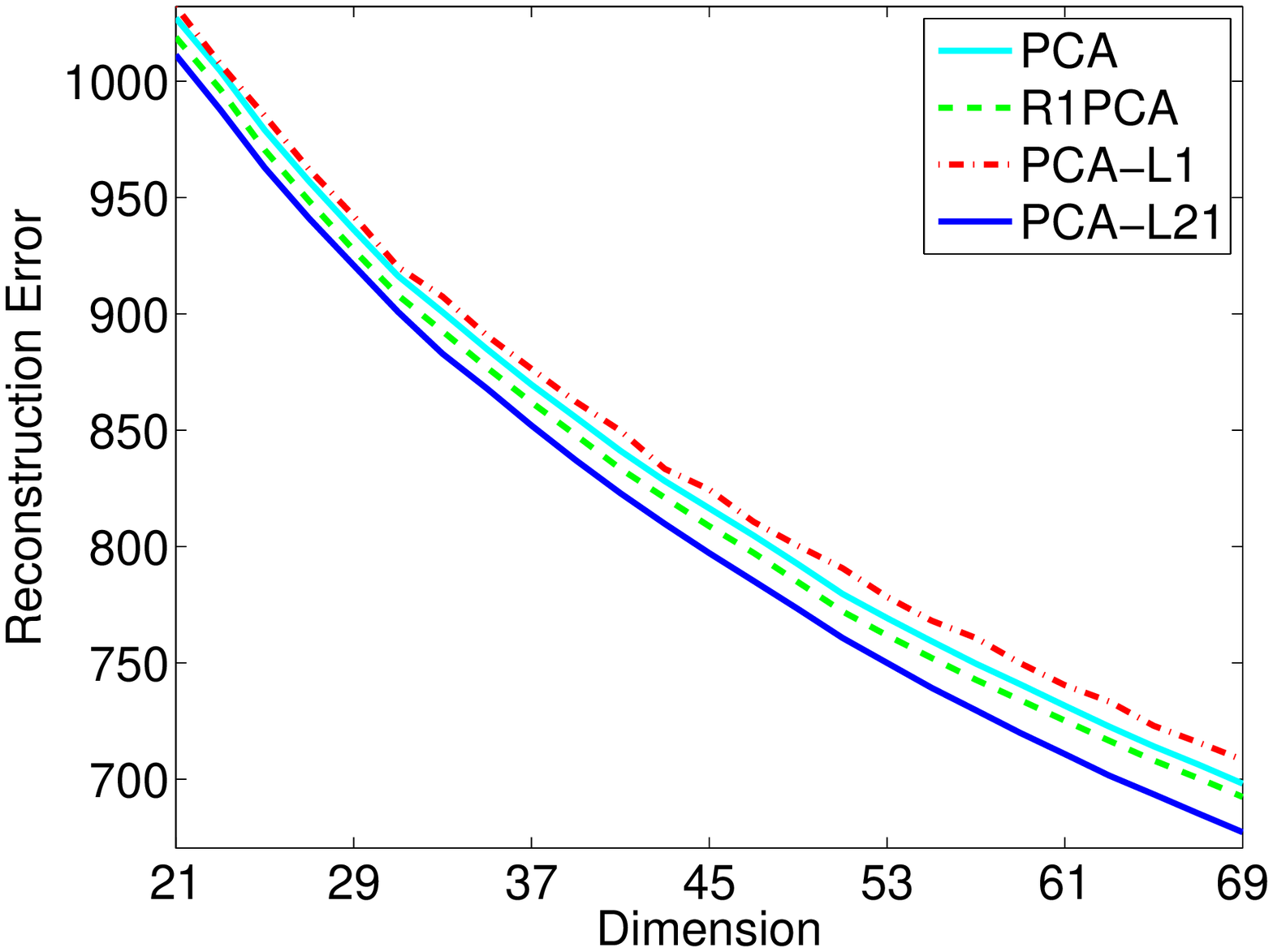}}
  \subfigure[Umist]{
    \label{waveform} 
    \includegraphics[height=4.cm]{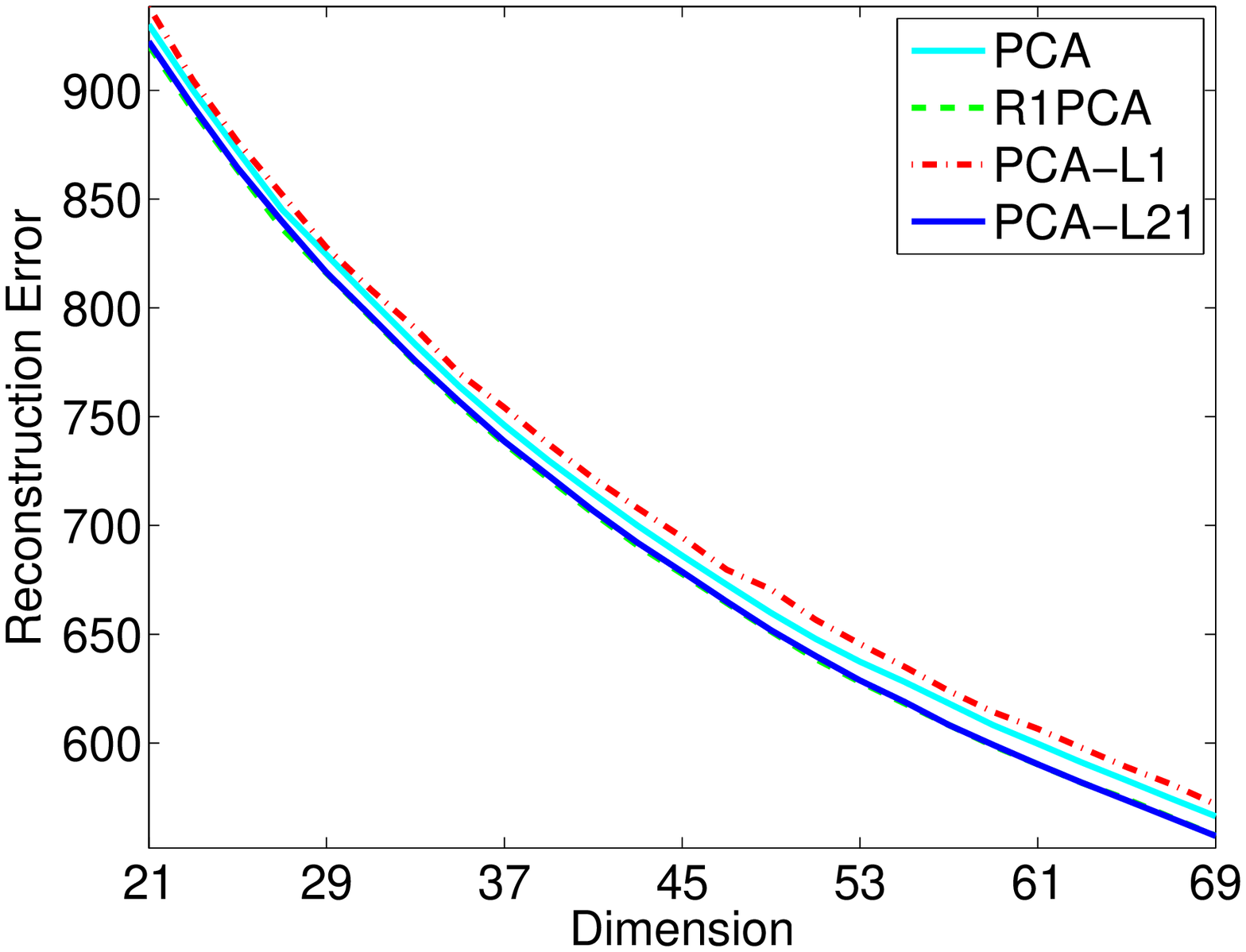}}
  \subfigure[AR]{
    \label{waveform} 
    \includegraphics[height=4.cm]{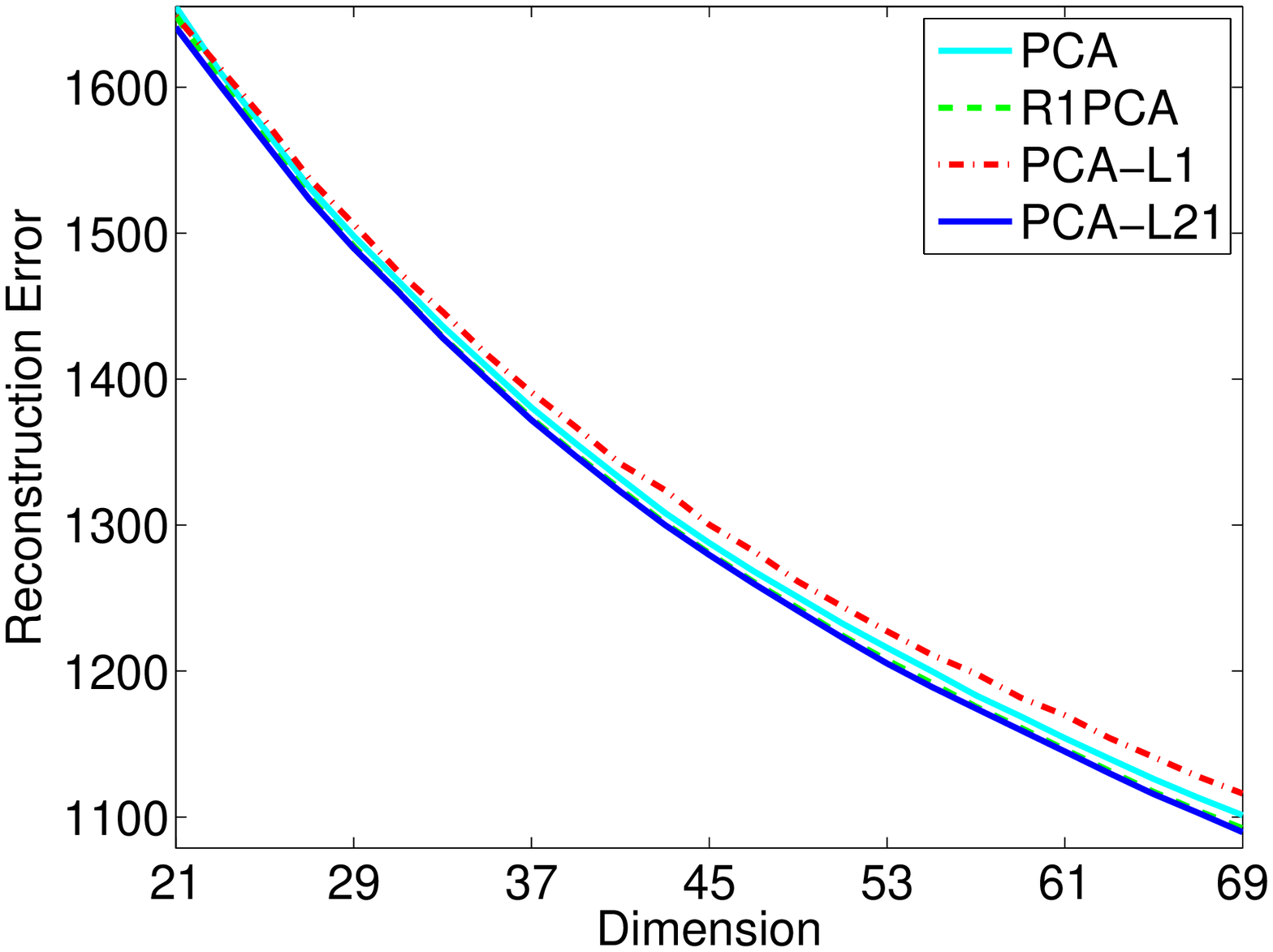}}
  \subfigure[XM2VTS]{
    \label{waveform} 
    \includegraphics[height=4.cm]{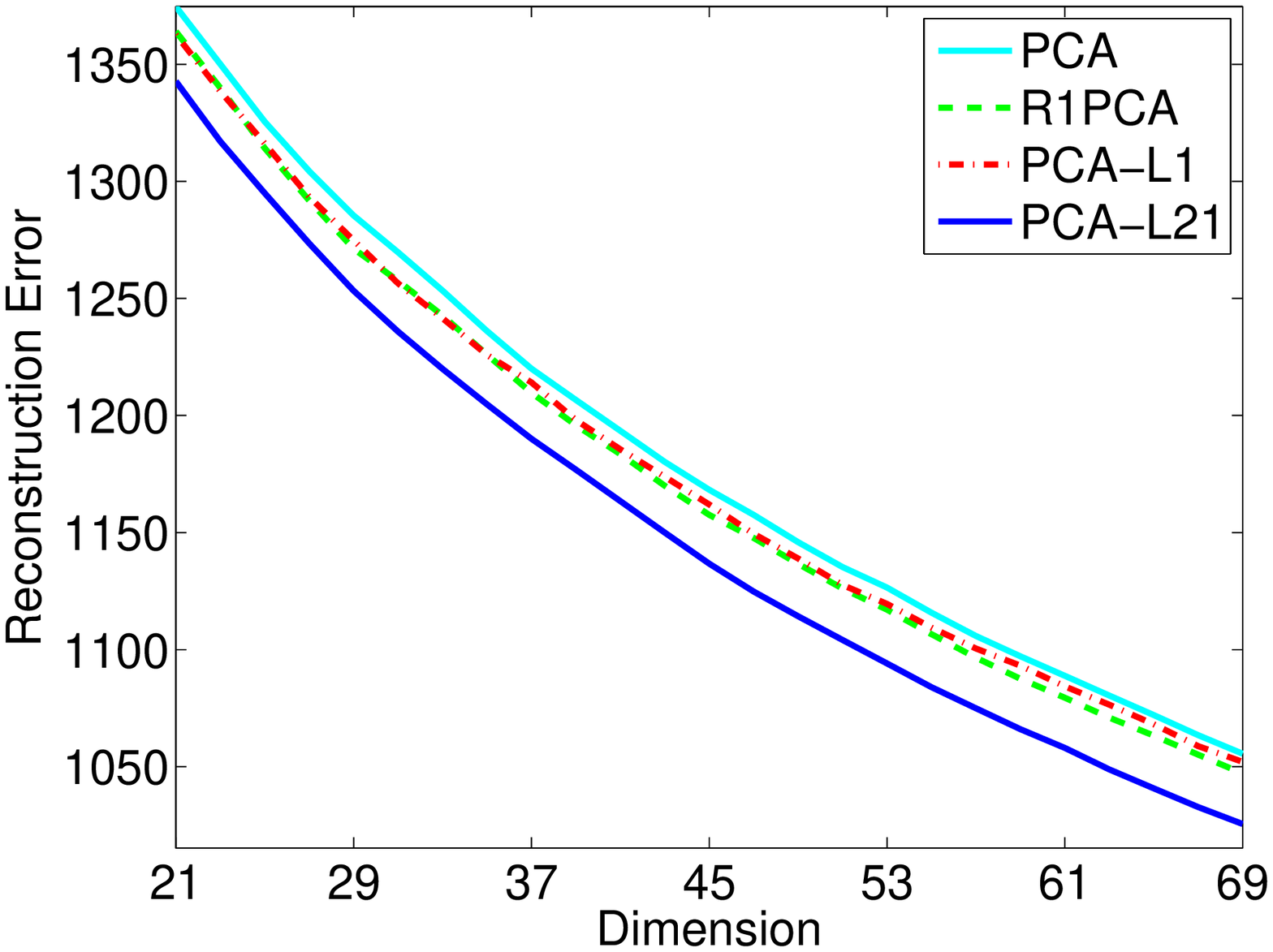}}
  \subfigure[Coil20]{
    \label{waveform} 
    \includegraphics[height=4.cm]{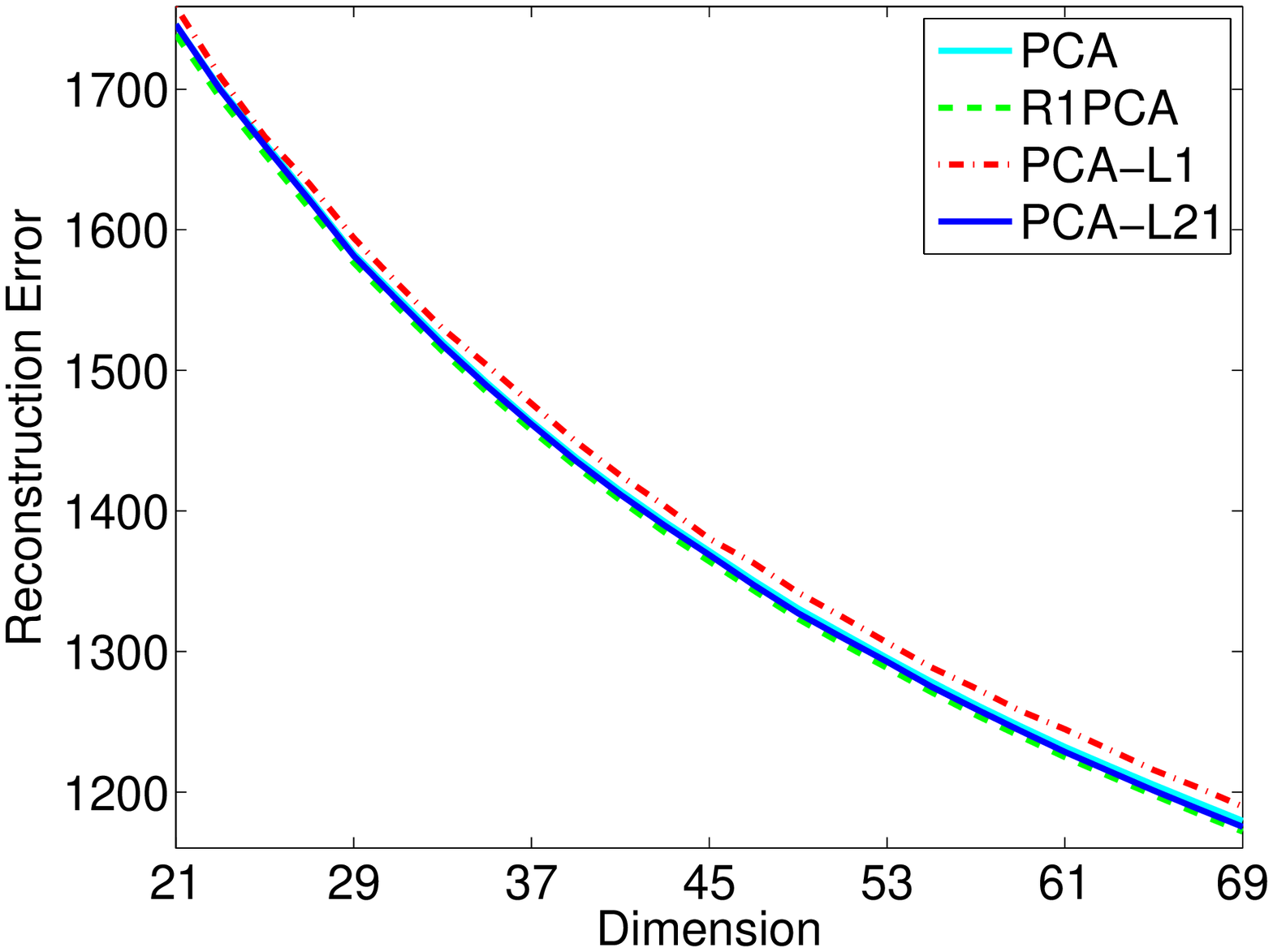}}
  \caption{Reconstruction errors (calculated by Eq.~(\ref{rce})) under different dimensions obtained by PCA, R1PCA, PCA-L1 and PCA-L21, respectively.
  In each data set, 20 percent images are randomly occluded with a randomly located square. }
  \label{objective2} 
\vskip 0.in
\end{figure*}

\begin{figure*}[!t]
\vskip 0.in
  \centering
  \subfigure[Yale]{
    \label{ionosphere} 
    \includegraphics[height=4.cm]{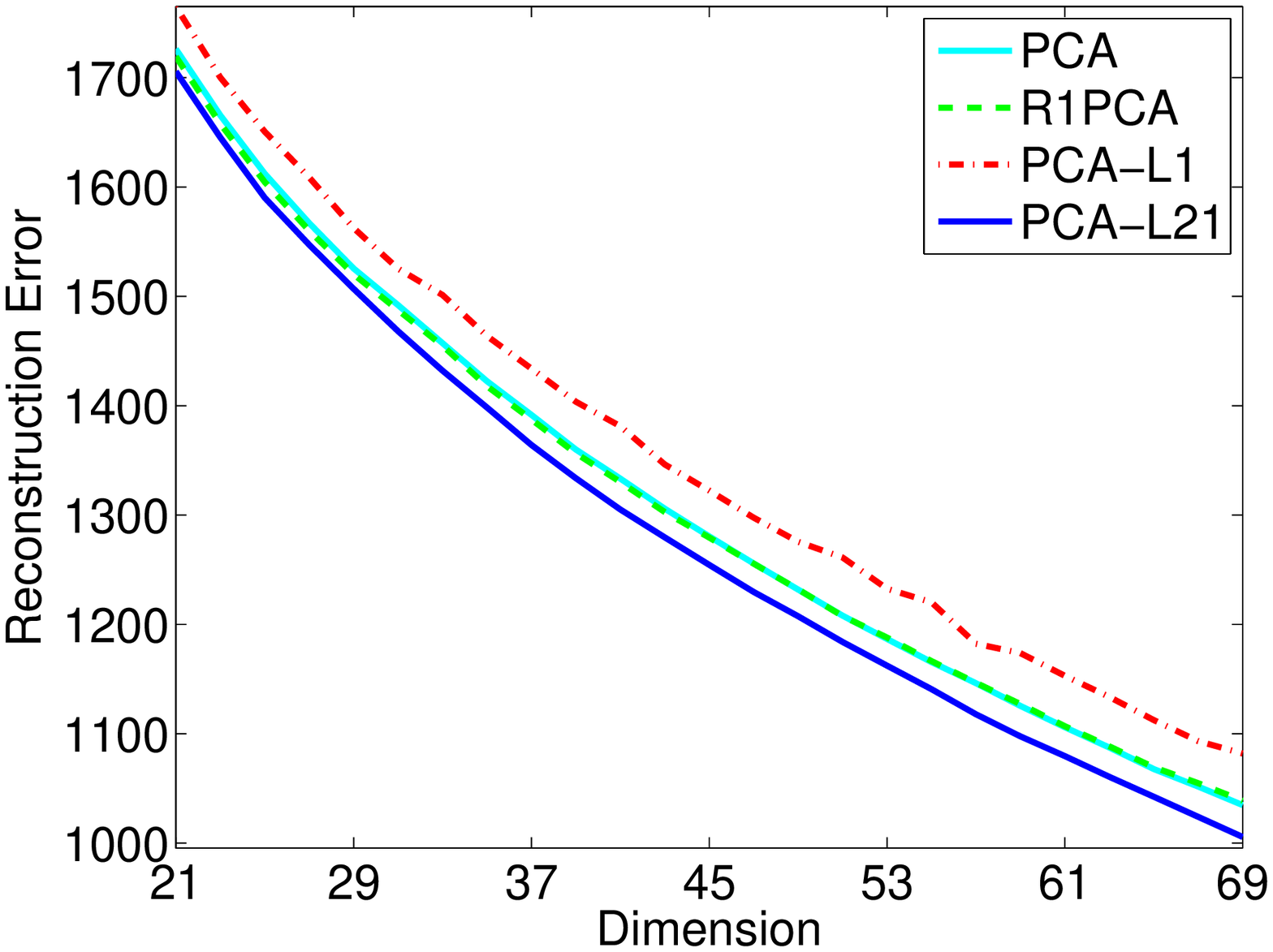}}
  \subfigure[AT\&T]{
    \label{waveform} 
    \includegraphics[height=4.cm]{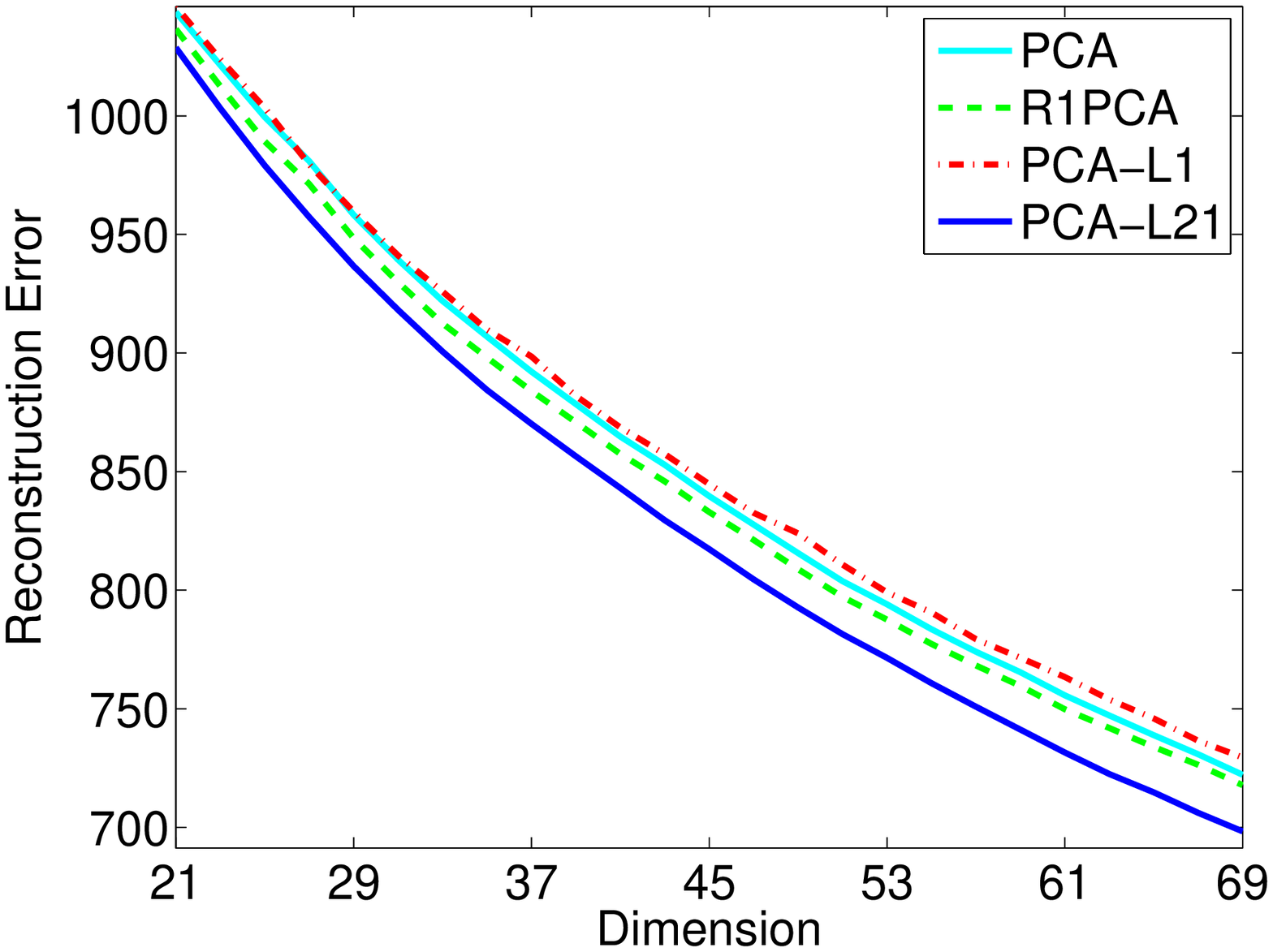}}
  \subfigure[Umist]{
    \label{waveform} 
    \includegraphics[height=4.cm]{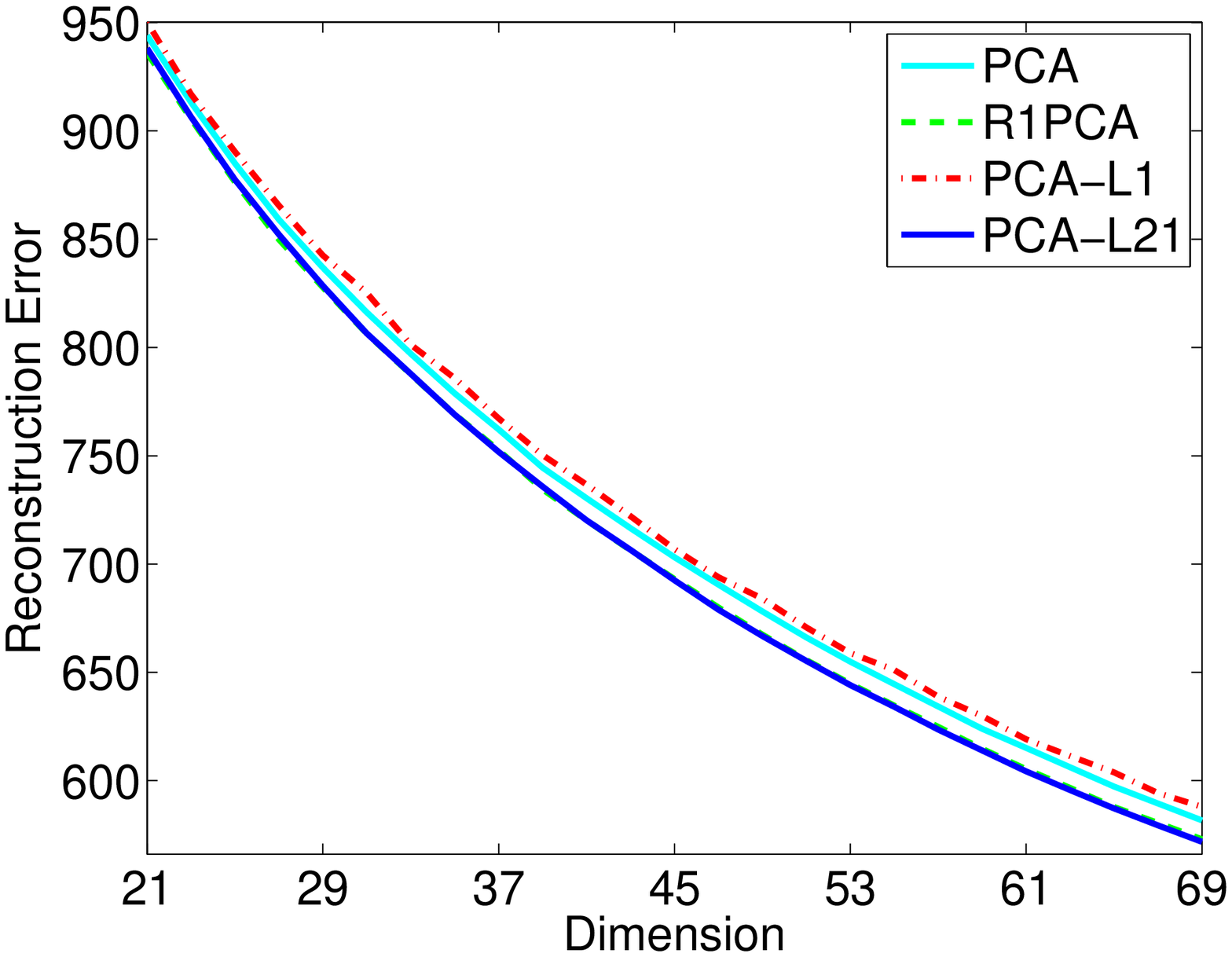}}
  \subfigure[AR]{
    \label{waveform} 
    \includegraphics[height=4.cm]{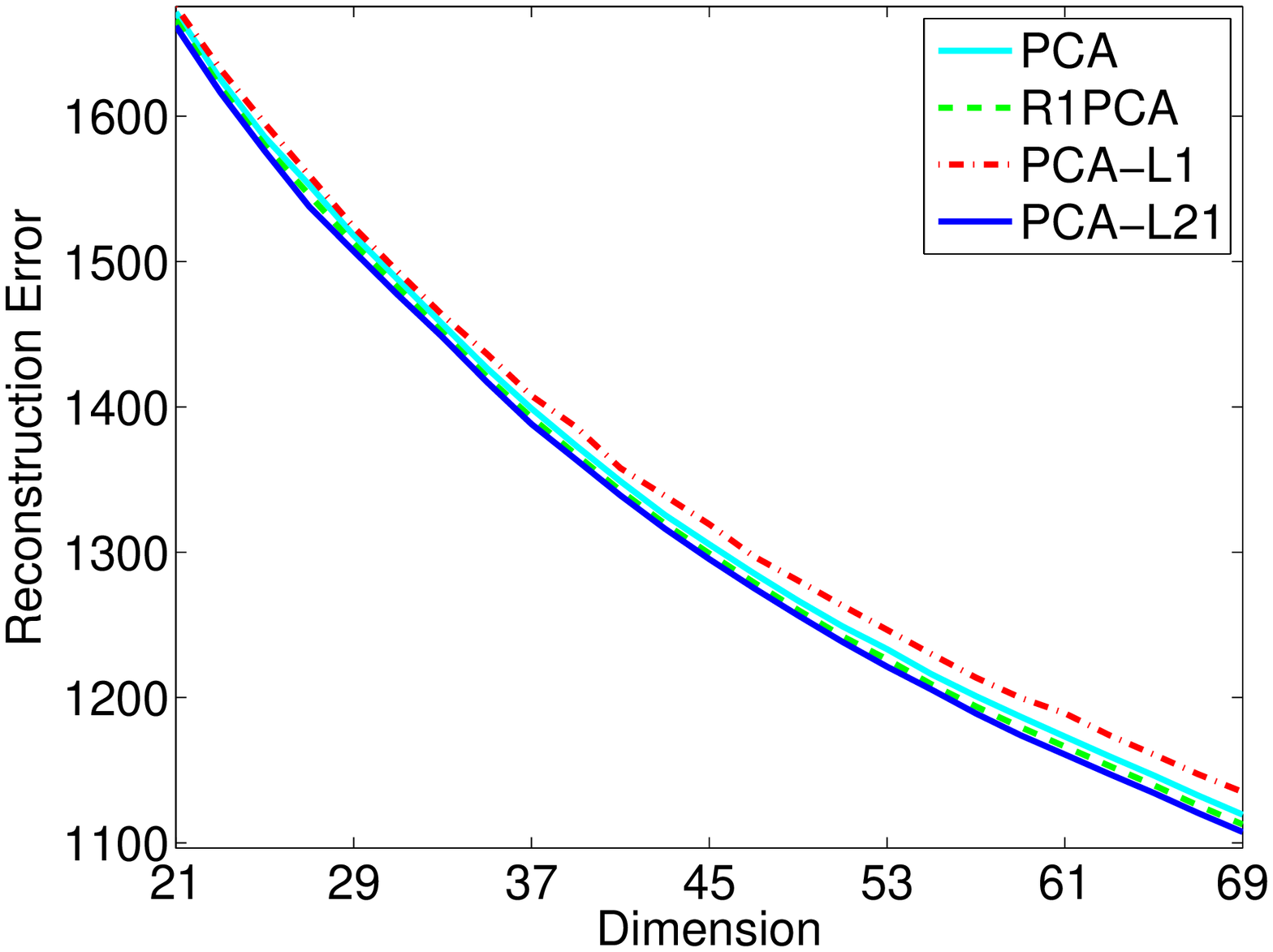}}
  \subfigure[XM2VTS]{
    \label{waveform} 
    \includegraphics[height=4.cm]{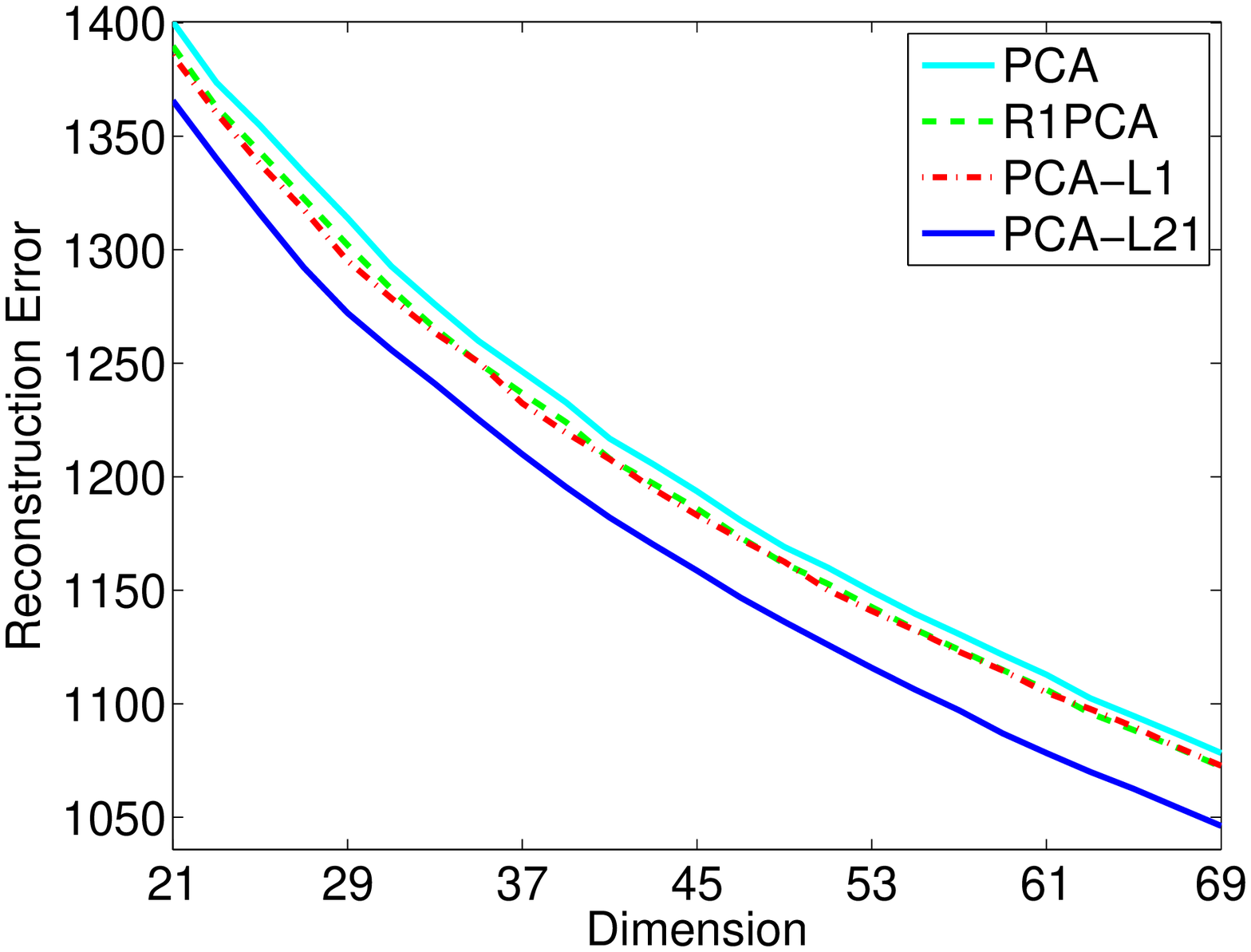}}
  \subfigure[Coil20]{
    \label{waveform} 
    \includegraphics[height=4.cm]{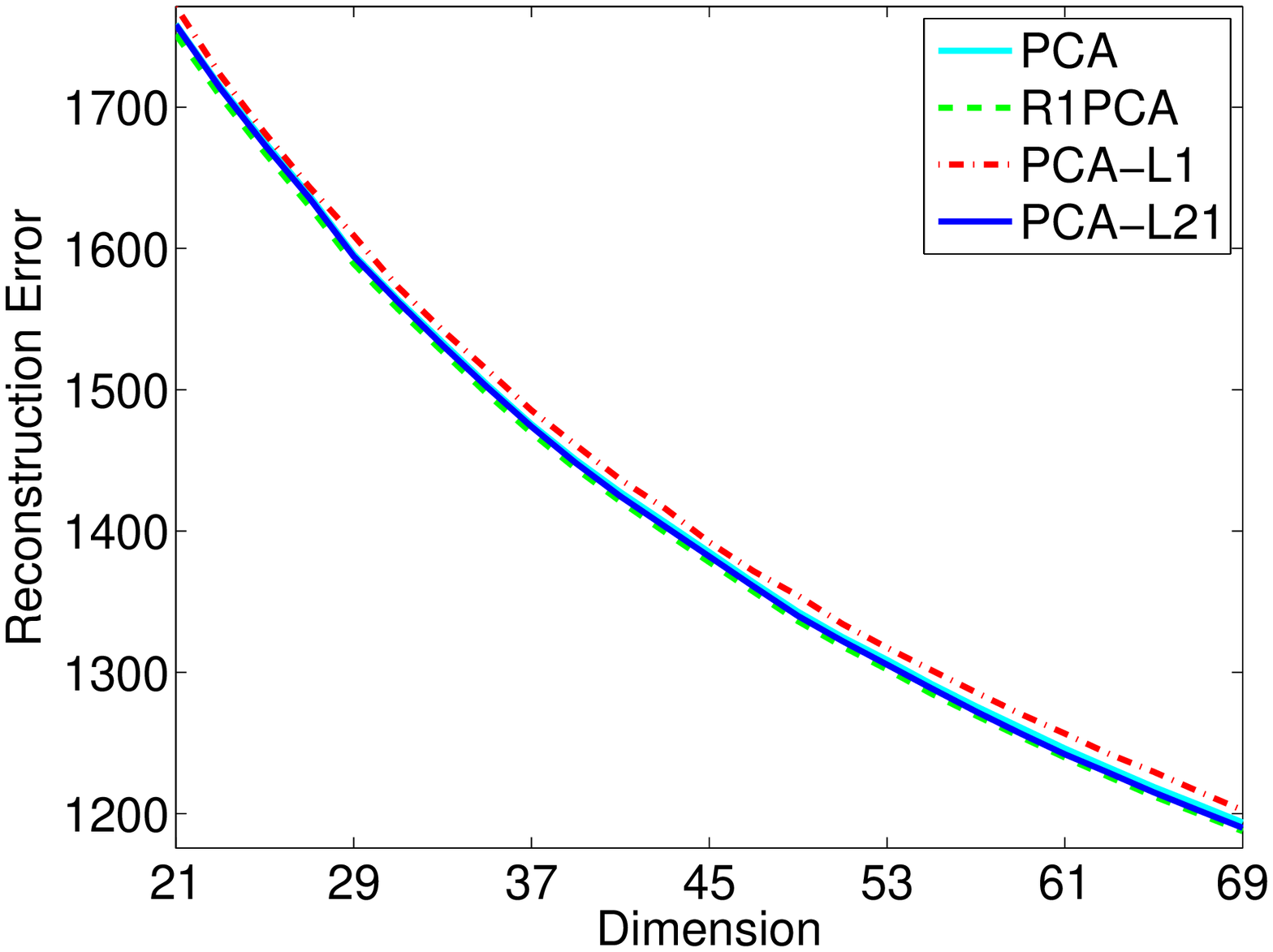}}
  \caption{Reconstruction errors (calculated by Eq.~(\ref{rce})) under different dimensions obtained by PCA, R1PCA, PCA-L1 and PCA-L21, respectively.
  In each data set, 30 percent images are randomly occluded with a randomly located square. }
  \label{objective3} 
\vskip 0.in
\end{figure*}

In the experiments, the projected dimension $m$ varies from 21 to 69. The results of the reconstruction error by PCA, R1-PCA, PCA-L1 and PCA-L21 are shown in Figures \ref{objective1}-\ref{objective3}. From the figures, we can see that PCA, R1-PCA and the proposed PCA-L21 are suitable for principal component analysis from the view of data reconstruction. When there are occlusions in the data, R1-PCA and PCA-L21 outperform PCA in terms of reconstruction error. We can also observed that in this experimental setting, PCA-L1 doesn't perform as good as PCA, which indicates that PCA-L1 is not closely connected to the minimization of reconstruction error, thus is not a good option for principal component analysis in some cases.

\subsection{Reconstruction errors with noise images}

\begin{figure}
\vskip 0.in
  \centering
\includegraphics[height=1.5cm]{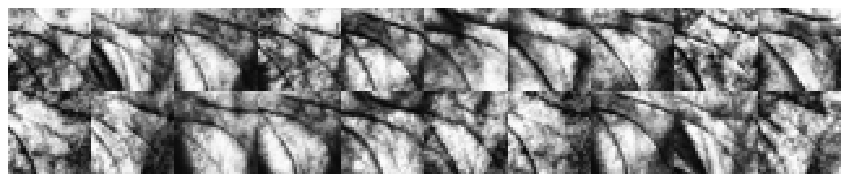}
\vskip -0.in
  \caption{Twenty image samples from the Palm image data set.}
\vskip -0.in
\label{datasamples1}
\end{figure}

In this experiment, two image data sets XM2VTS and Coil20 are used. For each data set, we add 10, 20, 30 percent images from the Palm image data set as the noise images, respectively. Some samples from the Palm data set are shown in Figure~\ref{datasamples1}.

We use the following reconstruction error to measure the quality of dimensionality reduction methods:
\begin{equation}
\label{rce1}
e(m) = \frac{1}{n} \sum\limits_{i = 1}^n {\left\| {x_i^{org}  - WW^T x_i^{org} } \right\|_2 },
\end{equation}
where $n$ is the number of training data (not including the noise image data from the Palm data set), $W \in
\Re^{d \times m}$ is the learned projection matrix by PCA, R1-PCA, PCA-L1 or PCA-L21,
$x_i^{org}$ is the $i$-th original training data (not including the noise image data from the Palm data set). Under this experimental setting, if a method is robust to the data outliers, its reconstruction error should be smaller than other non-robust methods.

\begin{figure*}[!t]
\vskip 0.in
  \centering
  \subfigure[XM2VTS, 10\% noise]{
    \label{ionosphere} 
    \includegraphics[height=4.cm]{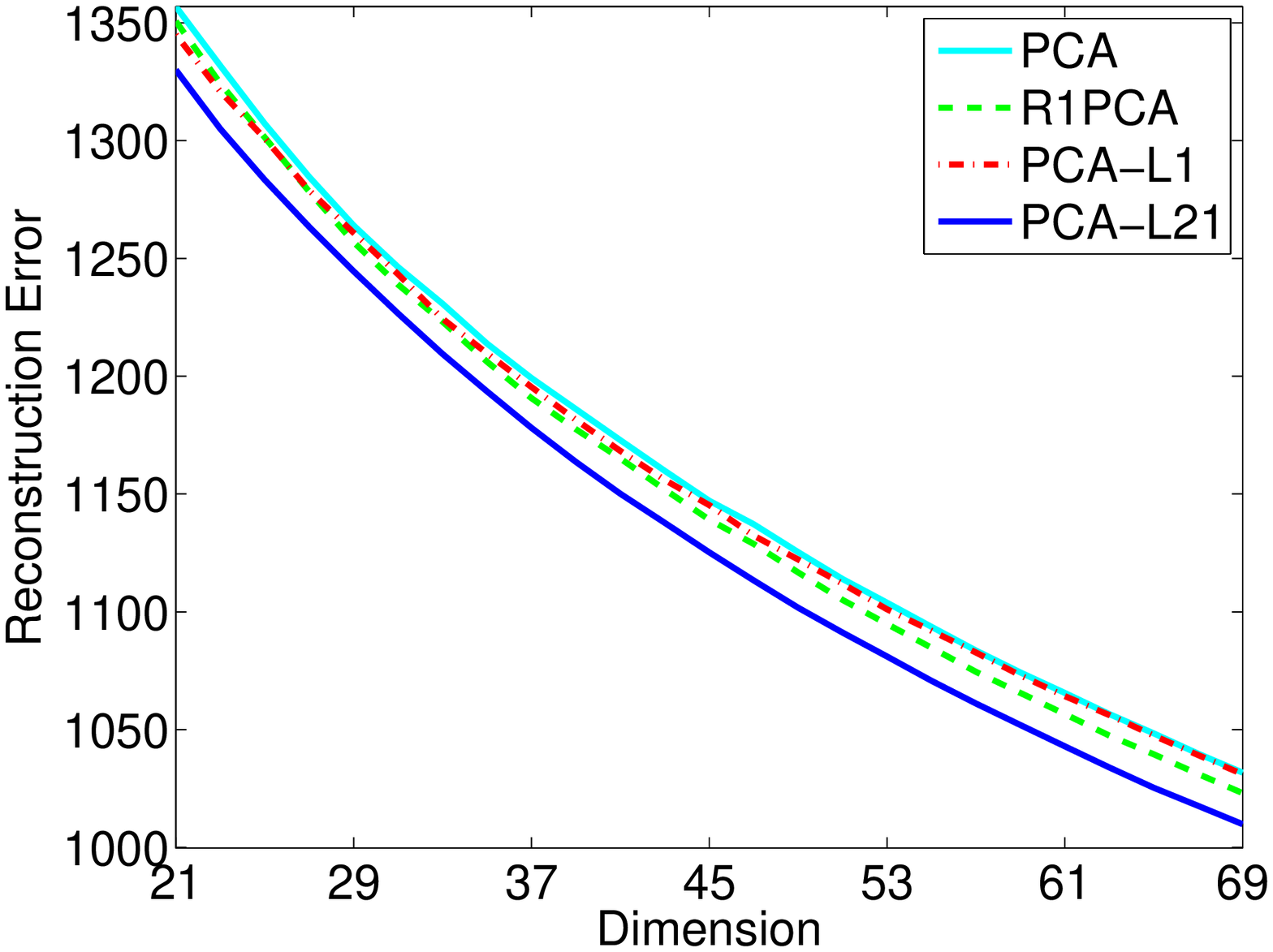}}
  \subfigure[XM2VTS, 20\% noise]{
    \label{waveform} 
    \includegraphics[height=4.cm]{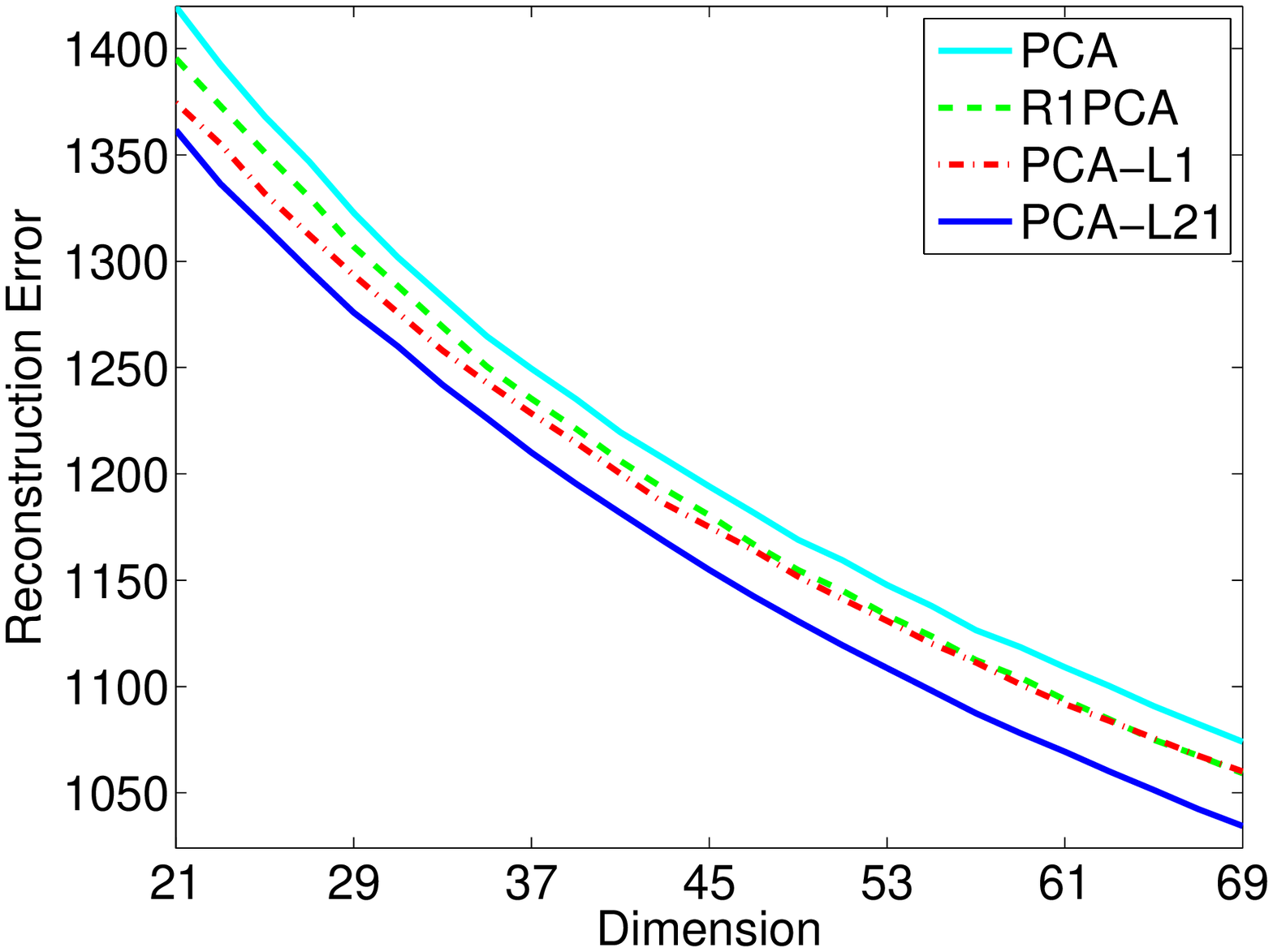}}
  \subfigure[XM2VTS, 30\% noise]{
    \label{waveform} 
    \includegraphics[height=4.cm]{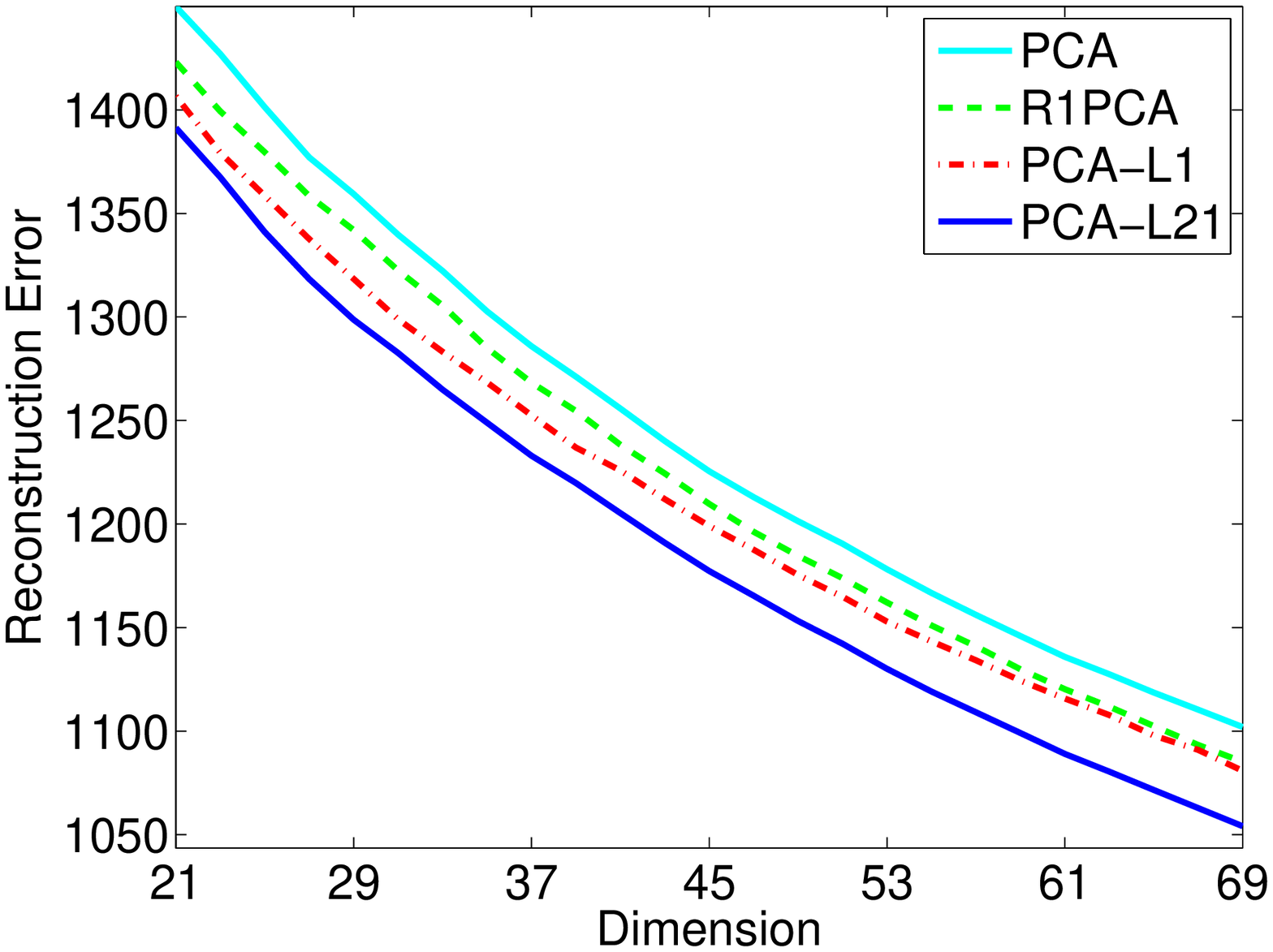}}
  \subfigure[Coil20, 10\% noise]{
    \label{waveform} 
    \includegraphics[height=4.cm]{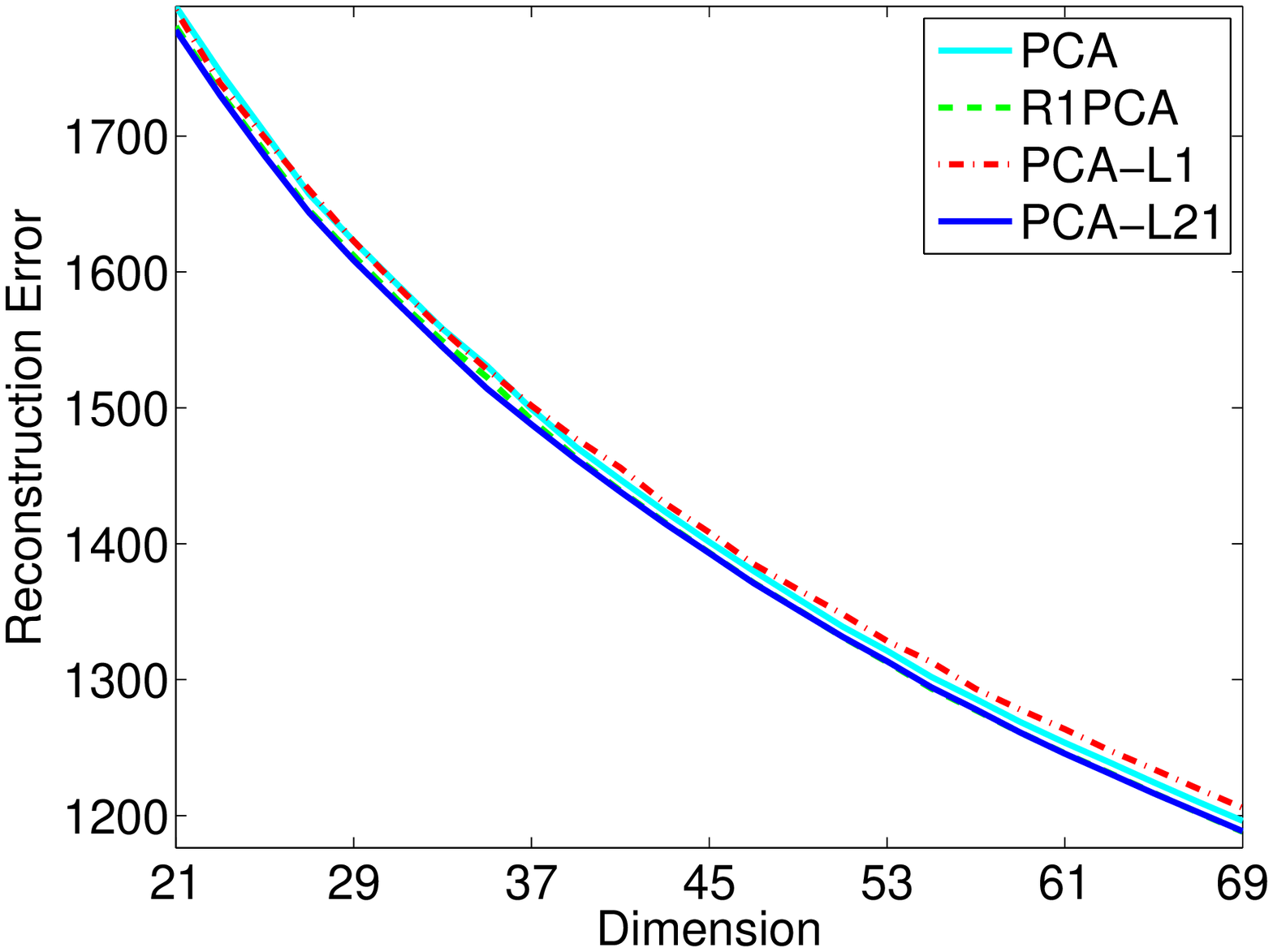}}
  \subfigure[Coil20, 20\% noise]{
    \label{waveform} 
    \includegraphics[height=4.cm]{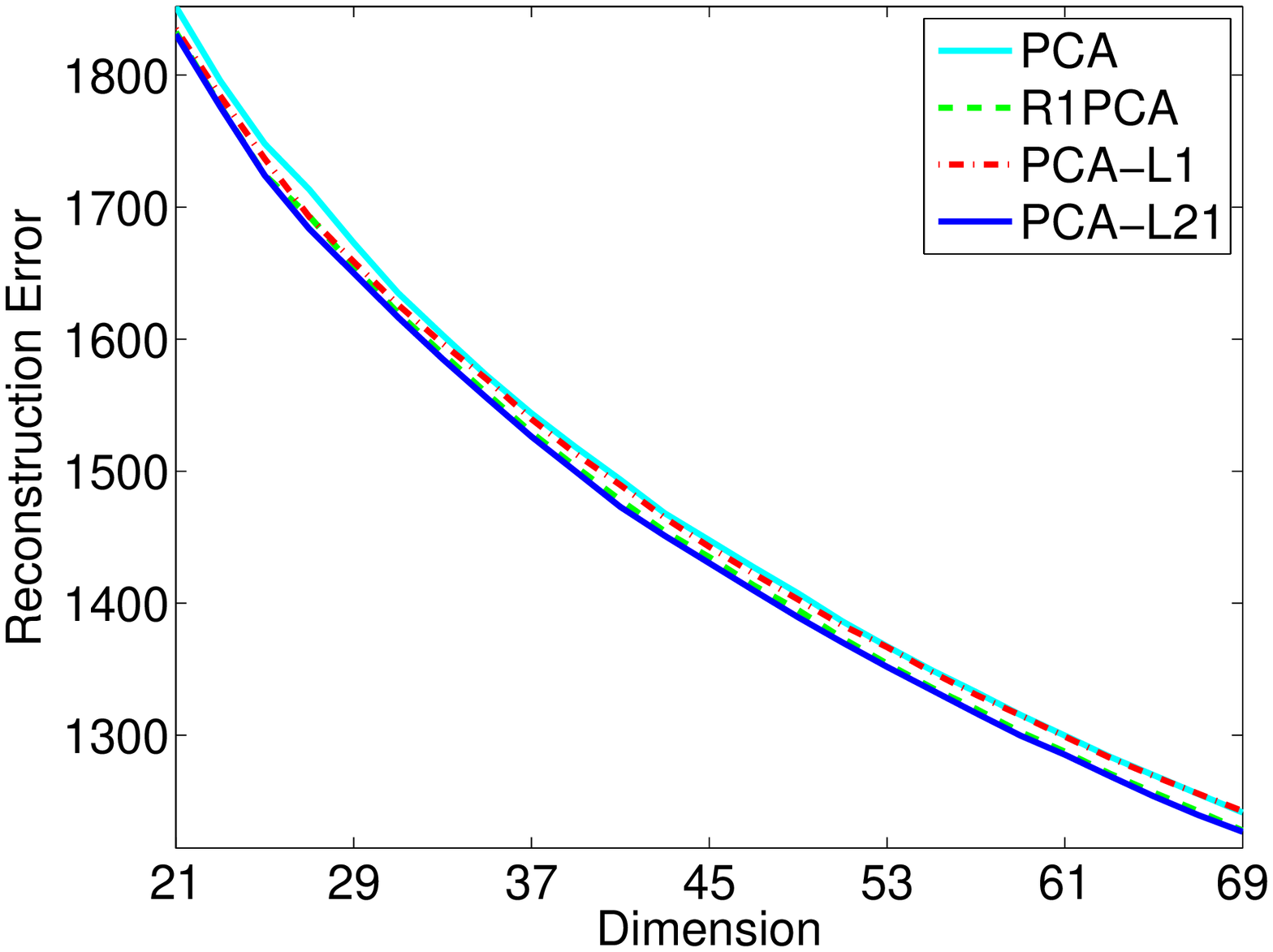}}
  \subfigure[Coil20, 30\% noise]{
    \label{waveform} 
    \includegraphics[height=4.cm]{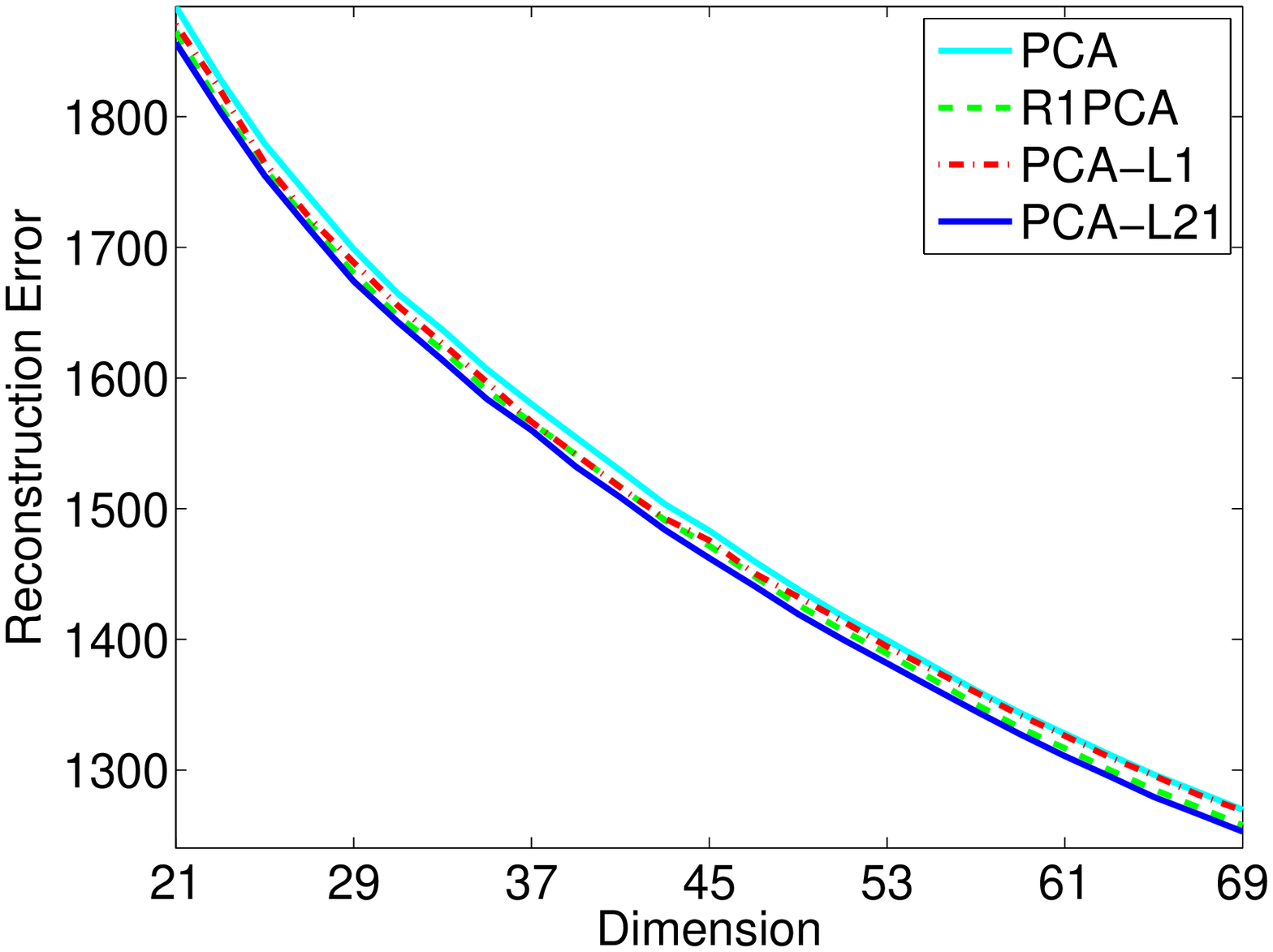}}
  \caption{Reconstruction errors (calculated by Eq.~(\ref{rce1})) under different dimensions obtained by PCA, R1PCA, PCA-L1 and PCA-L21, respectively.
  First row includes the results on the XM2VTS data set with adding 10, 20 or 30 percent images from the Palm data set as noise images.
  Second row shows the results on the Coil20 data set with adding 10, 20 or 30 percent images from the Palm data set as noise images.}
  \label{objective4} 
\vskip 0.in
\end{figure*}

In the experiments, the projected dimension $m$ varies from 21 to 69. The results of the reconstruction error by PCA, R1-PCA, PCA-L1 and PCA-L21 are shown in Figure \ref{objective4}.

We can see from the figures that, in this experimental setting, R1-PCA, PCA-L1 and the proposed PCA-L21 all outperform PCA in terms of reconstruction error, and our PCA-L21 consistently outperforms R1PCA, PCA-L1 and performs best in this case. The experimental results clearly indicates that
the proposed PCA-L21 is more suitable for principal component analysis than the traditional PCA when there are  outliers in the data.

\section{Conclusions}

A principal component analysis with L21-norm maximization was proposed in this paper. The L21-norm maximization based PCA is theoretically connected to the minimization of the reconstruction error, and thus is more suitable for principal component analysis than the L1-norm maximization based PCA proposed in \cite{RPCApami08}. To avoid the greedy strategy used in \cite{RPCApami08} for solving the L1-norm maximization problem, we propose an efficient optimization algorithm to solve a more general L21-norm maximization problem, which is non-greedy and is guaranteed to converge to a local solution. Moreover, we extend our algorithm to solve the more general maximization problem which can derive solutions for many related statistical learning models. Experimental results on real world data sets show that the proposed method is effective for principal component analysis, and always obtains smaller reconstruction error than the related methods under the same reduced dimension.


%
%

%

\section*{Acknowledgment}

This research was partially supported by NSF-IIS 1117965, NSF-CCF 0830780, NSF-DMS 0915228, NSF-CCF 0917274.



%

\bibliographystyle{IEEEtran}
\bibliography{elsart123}

%

%
%
%
%




\end{document}